\documentclass[letterpaper,10pt]{article}
\usepackage{indentfirst,latexsym,bm,amssymb,amsfonts,ifthen,color}
\usepackage{subfigure}
\usepackage{graphicx}
\usepackage{algorithm,algorithmic}
\usepackage{amsmath}
\usepackage{mathrsfs}
\usepackage{enumerate,enumitem}
\usepackage{multirow,diagbox}
\usepackage[left=1.1in,top=1in,right=1.1in,bottom=1in,letterpaper]{geometry}

\newcommand{\R}{\mathbb{R}}

\def\P{\mathcal{P}}

\def\M{\mathcal{M}}
\def\RR{\mathbb{R}}

\def\Rank{\mathrm{Rank}}

\usepackage{soul}
\setstcolor{red}
\setulcolor{blue}

\title{Manifold Based Low-rank Regularization for Image Restoration and Semi-supervised Learning
}
\author{Rongjie Lai
\thanks{Department of Mathematics, Rensselaer Polytechnic Institute, Troy, NY 12180,
         U.S.A. ({\tt lair@rpi.edu}). The research of Rongjie Lai is partially supported by NSF grant DMS--1522645.}
        \and Jia Li
        \thanks{Department of Mathematics, Rensselaer Polytechnic Institute, Troy, NY 12180,  ({\tt lij25@rpi.edu}).}
        }
\date{}

\begin{document}
\maketitle
\begin{abstract}
Low-rank structures play important role in recent advances of many problems in image science and data science. As a natural extension of low-rank structures for data with nonlinear structures, the concept of the low-dimensional manifold structure has been considered in many data processing problems. Inspired by this concept, we consider a manifold based low-rank regularization as a linear approximation of manifold dimension. This regularization is less restricted than the global low-rank regularization, and thus enjoy more flexibility to handle data with nonlinear structures. As applications, we demonstrate the proposed regularization to classical inverse problems in image sciences and data sciences including image inpainting, image super-resolution, X-ray computer tomography (CT) image reconstruction and semi-supervised learning. We conduct intensive numerical experiments in several image restoration problems and a semi-supervised learning problem of classifying handwritten digits using the MINST data. Our numerical tests demonstrate the effectiveness of the proposed methods and illustrate that the new regularization methods produce outstanding results by comparing with many existing methods.
\end{abstract}



\section{Introduction}
\label{sec:Intro}
Regularization methods play important roles in many ill-posed inverse problems arising in science and engineering. Examples include inverse problems considered in signal processing and image sciences such as image denoising, image impainting, image deconvolution~\cite{chan2005image,aubert2006mathematical}, just to name a few. Mathematically, a image restoration problem can be viewed as reconstructing a clean image $f$ from a degraded image $g$ based on the degradation relationship $\mathcal{D}(f) = g$. It is challenging to reconstruct $f$ from $g$ as the problem is usually ill-posed due to the highly underdetermined constraints and possible noise. Observations of natural image with prior information such as piecewise smoothness, shape edges, textures, repetitive patterns and sparse representations under certain transformations make regularization methods quite effective to handle image processing problems. Successful methods include the total variation (TV) methods, nonlocal methods and wavelet tight frame methods~\cite{ROF,buades2005non,Gilboa2008,Dong10mra-basedwavelet} and many others.
Moreover, regularization methods can also be considered in problems arising from data science. A typical example is semi-supervised learning, where tasks aim at labeling data from a small amount of labeled training data set. Regularization methods such as the harmonic extension method~\cite{zhu2003semi} have been considered to this type of ill-posed problem. In this paper, we consider a different regularization, called manifold based low-rank (MLR) regularization as a linearization of manifold dimension, which generalizes the global low-rank prior knowledge for linear objects to manifold-region-based locally low-rank for nonlinear objects.

The idea of the MLR proposed in this paper is inspired by a recent method called the low-dimensional manifold model (LDMM) discussed in~\cite{osher2016low}. Using the image patches discussed in nonlocal methods~\cite{buades2005non,peyre2009manifold}, the LDMM interprets image patches as a point cloud sampled in a low-dimensional manifold embedded in a high dimensional ambient space, which provides a new way of regularization by minimizing the dimension of the corresponding image patch manifold. This can be explained as a natural extension of the idea of low-rank regularization for linear objects to data with more complicated structures. Moreover, the authors in~\cite{osher2016low} elegantly find that the point-wisely defined manifold dimension can be computed as a Dirichlet energy of the coordinate functions on the manifold, whose corresponding boundary value problem can be further solved by a point integral method proposed in~\cite{li2014point}. The LDMM performs very well in image inpainting and super-resolution. This model is later considered in collaborative ranking problems~\cite{kuang2016harmonic}. Based on weighted graph laplacian (WGL), an improvement of LDMM called LDMM+WGL is proposed more recently in \cite{shilow}.

\begin{figure}[t]
\centering
\begin{tabular}{cc}
\includegraphics[width=0.56\linewidth]{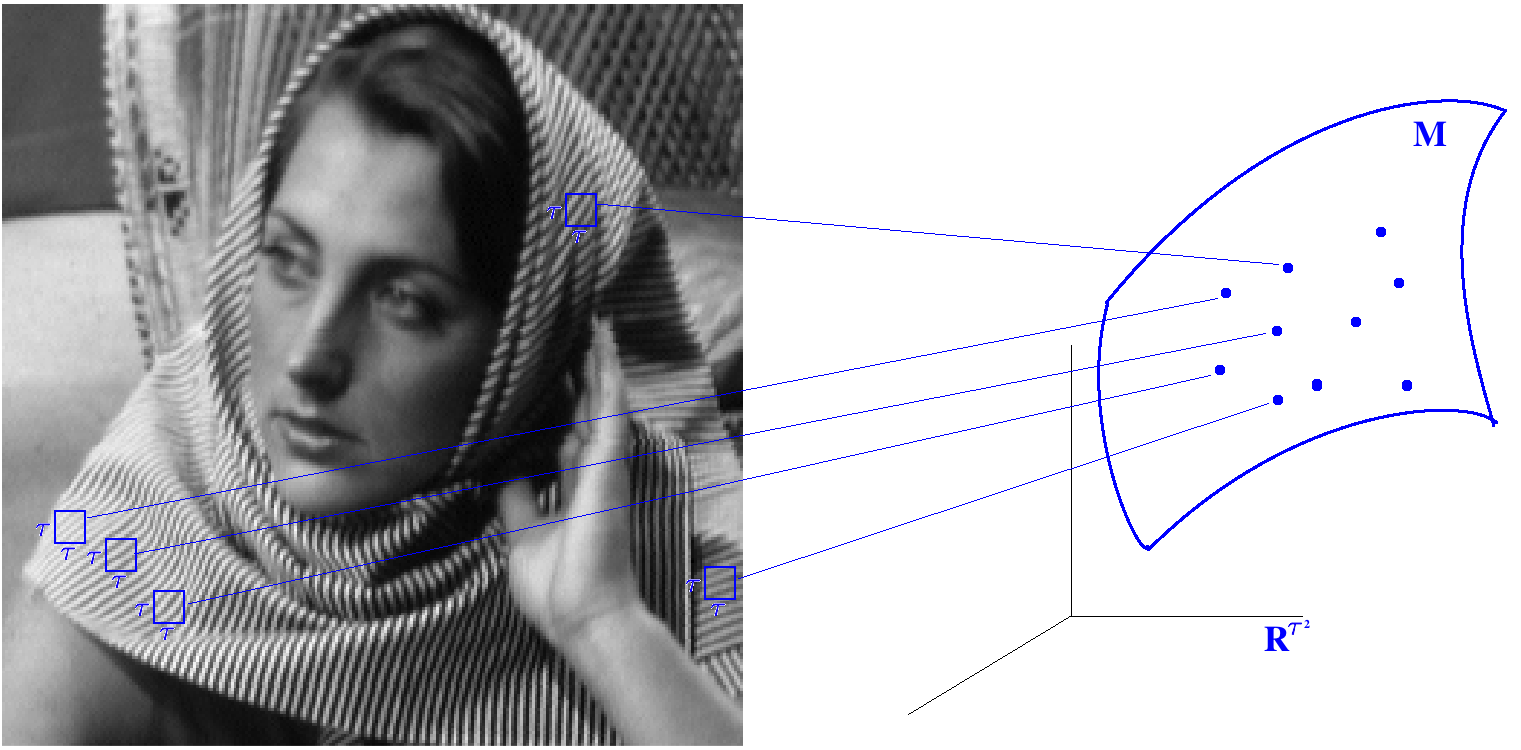} &
\includegraphics[width=0.31\linewidth]{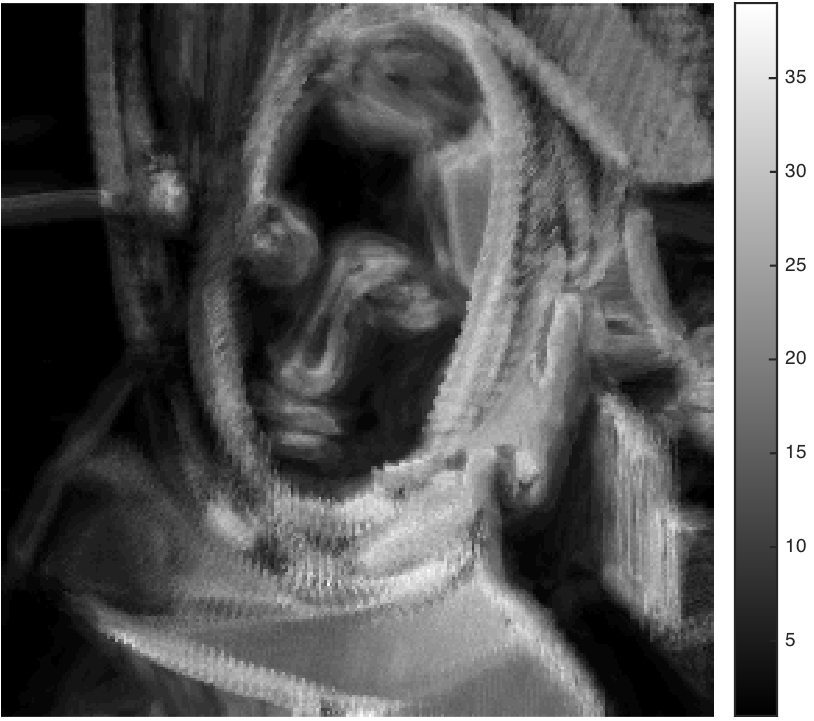} \\
\end{tabular}
\caption{Left: A clean Barbara image and the corresponding patch manifold. Right: The point-wise rank function $\Rank(R_{\mathcal{M},x})$ of the patch manifold with patch size 11 $\times$ 11.}
\label{fig:Patch Manifold}
\end{figure}

In this paper, instead of representing the manifold dimension as a manifold-derivative involved quantity~\cite{osher2016low}, we propose a linear approximation of the manifold dimension. Note that the quantity of the dimension at each point $x\in\mathcal{M}$ is the same as the dimension of the tangent space at $x$. This quantity only depends on a local neighborhood of $x$ on $\mathcal{M}$, which can be approximated as the rank of the covariance matrix generated by the set of $K$-nearest-neighbourhood (KNN) points of $x$ on $\M\subset\RR^n$ in the discretized sense. In other words, the low-dimensional property of $\M$ at $x$ can be linearly approximated as the low-rank property of the this corresponding covariance matrix, which is essentially the same as the low-rank property of the matrix $R_{\mathcal{M},x}$ formed by those KNN points near $x$.
As an example illustrated in Figure \ref{fig:Patch Manifold}, we construct a patch manifold of the Barbara image using patch size $11 \times 11$. This leads to a set of image patches represented as a point clouds in $\RR^{121}$. The rank of $R_{\mathcal{M},x}$ for the Barbara image is color-coded in the right image of Figure \ref{fig:Patch Manifold}, which clearly illustrates that $\Rank(R_{\mathcal{M},x})$ has low value for this natural image.
As a linear approximation of the $Dim_{\mathcal{M}}(x)$ proposed in \cite{osher2016low}, the manifold based quantity $\Rank(R_{\mathcal{M},x})$ does not involve with any manifold differential operators, which has potential to apply this concept to more general data processing problem such as a preliminary example demonstrate in section~\ref{sec:SemiSuperLearning}. On the other hand, this consideration is reasonable as the globally defined ``Rank" can only handle linear objects, while this manifold based locally defined $\Rank$ has advantages to regularize data with nonlinear structures.

Based on the MLR prior knowledge, we use the matrix nuclear norm relaxation for matrix rank as the method considered in low-rank matrix completion theory~\cite{CandesRe2008} and apply MLR to the image patch manifold for image restoration problems including image inpainting, image super-resolution and X-ray computer tomography (CT) image reconstruction. It is clear the definition MLR relies on the construction of KNN which is essentially dependent on the manifold structure. Therefore, a split-Bregman method~\cite{goldstein2009split} is considered to solve the proposed model by iteratively updating the manifold structure and the objective image. Moreover, we also apply the proposed regularization for a semi-supervised learning problem, where MLR is applied to a labeling matrix with a fixed manifold structure provided by the input data. Our numerical results tested for a benchmark data set of handwritten digits illustrate the effectiveness of the proposed method.

The rest part of this paper is organized as follows. In Section \ref{sec:MLRimage}, we discuss our manifold based low-rank regularization for the image restoration problems including image inpainting, image super-resolution and X-ray CT image reconstruction. Detailed models and numerical algorithms for various image processing problems are discussed. In Section \ref{sec:SemiSuperLearning}, we consider the manifold based low-rank model to a semi-supervised learning problem. Intensive numerical experiments and comparisons with existing methods are conducted in Section \ref{sec:experiments}. We conclude our work in Section \ref{sec:conclusions}.

\section{Manifold based low-rank regularization for image restoration}
\label{sec:MLRimage}
In this section, we consider the MLR method for image restoration problems including image inpainting, image super-resolution and X-ray CT image reconstruction. The idea of MLR is applied to a image patch manifold with a fixed patch size similar as the way proposed in~\cite{osher2016low}. We further relax the problem of matrix rank minimization as a problem of matrix nuclear norm optimization and solve the proposed optimization problem based on the split Bregman iteration~\cite{goldstein2009split} and the singular value thresholding algorithm~\cite{cai2010singular}.

The classical image restoration models mainly focus on local properties of the objective image such as smoothes and jumps. Image features can be further enhanced due to its possible repetitive patterns non-locally. The nonlocal based image restoration methods \cite{buades2005non,dabov2006image,Gilboa2008} extract and match non-local repetitive structures of images using image patches. Given a discrete image $f \in \RR^{m \times n}$ defined on a domain $\mathscr{I} =  \{1,2,\ldots,m\} \times \{1,2,\ldots,n\}$, a size $\tau = 2 \eta +1$ patch transform $\mathcal{P}$ can be defined by:
\begin{equation}\label{patch minifold}
\begin{aligned}
\mathcal{P}: \RR^{m \times n}  & \rightarrow \RR^{\tau^2 \times mn}  \\ \quad f & \mapsto \mathcal{P}(f) ,\qquad
\mathcal{P}(f) (s,x)   = \tilde{f}(x+s), ~s \in \mathscr{P},x \in \mathscr{I},
\end{aligned}
\end{equation}
where $x$ is the center of each patch, $\mathscr{P} =  \{-\eta, -\eta+1, \ldots,0,1,\ldots,\eta-1,\eta\}^2$ represents the patch index set and $\tilde{f} \in \R^{(m+ 2 \eta) \times (n+ 2 \eta)}$ is a proper extension (symmetric extension in this paper) of $f$ such that $\tilde{f}(x) = f(x), \forall x \in \mathscr{I}$.
An essential observation of nonlocal methods is that images can be restored by enhancing similar patterns which may not lie in nearby regions of $\mathscr{I}$ domain. Therefore, comparing with the direct regularization methods on the image domain of $f$, the quality of image restoration can be usually improved using nonlocal methods. For instance, nonlocal based variational methods~\cite{buades2005non,Gilboa2008,zhang2010wavelet} and nonlocal based wavelet frame based methods~\cite{quan2015data} demonstrate outstanding image restoration results.

Given a patch matrix $\mathcal{P}(f)$, one can regard each patch $\mathcal{P}(f) (\cdot,x)$ as a $\tau^2$ dimensional column vector. Consequently, $\mathcal{P}(f)$ can be viewed as a set of points in $\R^{\tau^2}$. To conduct further analysis of this point cloud, we model $\mathcal{P}(f)$ as a set of points sampled on a manifold $\mathcal{M} \in \R^{\tau^2}$. Thereafter, we also abuse the notation $x$ as the corresponding point  $\mathcal{P}(f) (\cdot,x)$ on $\mathcal{M}$. This manifold interpretation has been proposed in existing work~\cite{peyre2009manifold,osher2016low}. More recently, \cite{osher2016low} proposes a low dimensional manifold model (LDMM) for image restoration. This work observes that the dimension of patch manifold $\M$ should intrinsically have a low-dimensional structure and proposes to regularize the dimension of the patch manifold $\mathcal{M}$ for image restoration. Moreover, the authors elegantly show that the dimension function $Dim(\mathcal{M})$ at $x \in \mathcal{M}$ can be represented by $Dim_{\mathcal{M}}(x) =\sum_{1 \leq s \leq \tau^2}\|\nabla_\mathcal{M} (\mathcal{P}(f)(s,\cdot))(x)\|_2^2$, which transforms the dimension regularization problem to be a variational partial differential equation model that is proposed to solve using a point integral method discussed in~\cite{li2014point}.
Later on, \cite{kuang2016harmonic} generalized the LDMM model into matrix completions with better performance than traditional low-rank regularized model in completing the Netflix matrix \cite{bennett2007netflix} which does not have exactly global low rank.

\subsection{Manifold based low-rank regularization for the patch manifold}
Inspired by the regularization of the manifold dimension represented as a manifold derivative involved quantity~\cite{osher2016low}, we propose a linear approximation of the manifold dimension in the following way. Note that the quantity of dimension at each point $x\in\mathcal{M}$ is the same as the dimension of the tangent space $\mathcal{T}_{x}\mathcal{M}$ at $x$ which only relies on a local neighborhood of $x$ on $\mathcal{M}$. In the discrete sense of $\M$ sampled as the patch matrix $\P(f)$, the quantity $\mathrm{dim}(\mathcal{T}_x\M)$ can be approximated as the rank of the covariance matrix generated by the set of $K$-nearest-neighbourhood (KNN) points of $x$ in $\P(f)$. In other words, the low-dimensional property of $\M$ at $x$ can be linearly approximated as the low-rank property of the corresponding covariance matrix, which is essentially the same as the low-rank property of the matrix formed by those KNN points near $x$. More precisely, if we define the restriction operator $R_{\mathcal{M},x}$ as the KNN points near $x$, then the low-dimensional prior knowledge of the patch manifold $\M$ at $x$ can be linearly approximated as the low-rank prior knowledge of the matrix formed by points in $R_{\M,x}$ denoted as $R_{\M,x}(\P(f))$. Namely, we define the manifold based rank at $x$ as $\text{Rank}_\M(x) = \text{Rank}(R_{\mathcal{M},x} (\mathcal{P}(f)))$.

For image restoration problems, if the fidelity information is $\mathcal{D}(f) = g$ as a constraint where $\mathcal{D}$ is a degradation operator, by regularizing $\text{Rank}_\M(x)$ for all the point $x$, we consider the following the manifold based low-rank regularization for image restoration:
\begin{equation}\label{image restoration}
\min_{\mathcal{M} \subset \R^{\tau^2} , f} ~\sum_{x \in \mathscr{I}} \Rank(R_{\mathcal{M},x} (\mathcal{P}(f)), \quad \text{s.t.}\quad \mathcal{P}(f) \subset \mathcal{M}, \quad \mathcal{D}(f) = g,
\end{equation}
On the one hand, the minimization of the rank, or the $\ell_0$ norm of the singular value, is NP hard to  be optimized generally.  Therefore,  $\ell_1$ norm of the singular value, or the nuclear norm of the localized matrix, is an appropriate way to relax the local rank as the pioneer work of low-rank matrix completion theory developed in \cite{CandesRe2008}. The minimization of nuclear norm can be solved by applying the singular value thresholding (SVT) algorithm \cite{cai2010singular}. On the other hand, we observe that it is necessary to smoothen the images, or enhance the features and textures in practice. Therefore, one can apply some positive/negative diffusion based regularization \cite{gilboa2002forward} to $f$ to guarantee the smoothness of the object image. For example, we choose the diffusion term as the non-local gradient operator defined in \eqref{non local gradient def}.
\begin{equation}\label{non local gradient def}
(\nabla_\mathcal{M} f)(x,y):=(f(y)-f(x))\sqrt{\omega(x,y)}, \qquad x,y \in \mathscr{I}.
\end{equation}
Therefore, a MLR image restoration model can be stated as:
\begin{equation}\label{non local gradient}
\min_{\mathcal{M} \subset \R^{\tau^2} , f} ~\sum_{x \in \mathscr{I}}  \|R_{\mathcal{M},x} (\mathcal{P}(f)) \|_* +\frac{\lambda}{2}\|\nabla_\mathcal{M} f\|_2^2, \quad \text{s.t.}\quad \mathcal{P}(f) \subset \mathcal{M}, \quad \mathcal{D}(f) = g,
\end{equation}
when $\lambda>0$ the regularization term $\frac{\lambda}{2}\|\nabla_\mathcal{M} f\|_2^2$ represents a diffusion term which can smoothen the regions. When $\lambda<0$, the regularization represents inverse diffusion which can enhance the patterns \cite{gilboa2002forward,gilboa2004,buades2006image}. Otherwise, $\lambda=0$ leads the model \eqref{non local gradient} identical to model \eqref{image restoration} as a pure MLR regularized image restoration model.

We remark that a close related work~\cite{dong2014compressive} imposed the low-rank regularization in a non-local transform domain of images, which is applicable to recover images from missing Fourier coefficients. In particular, to improve the robustness of the algorithm, the low-rank regularization is considered to the grouped ``similar patches", which can be regarded as a type of ``locally low-rank regularization" although \cite{dong2014compressive} did not explicitly view the ``low-rank" in manifold sense. In addition, this method considers to group patches without sufficient overlapping, thus it only includes a rough sampling on the patch manifold which may not be able to accurately reflect the low-dimensional structure of the patch manifold. 

\subsection{MLR for image inpainting}
Image inpainting \cite{BSCB} is a process to restore images whose pixels are missing, over-written or corrupted. More precisely, the inpainting problem aims at reconstructing an image $f$ only based on its partial information on a given set $\Omega \subset \mathscr{I}$. Such ill-posed problem is generally based on some assumptions such that the object image $f$ is piecewise smooth, or has repetitive textures. With these assumptions, regularization inpainting methods, such as variational PDE based models \cite{chan2001nontexture,shen2002mathematical, shi2016weighted}, wavelet based models \cite{CSZ,chan2005frameletreport,cai2008framelet,cai2008simultaneous,dong2012wavelet} and low dimensional manifold model \cite{osher2016low} have been proposed.

We would like to demonstrate that MLR model can restore images and preserve both the piecewise smooth regions and textures from a small random portion of information. As a special case of \eqref{non local gradient}, the low-rank regularized image inpainting model can be stated as:
\begin{equation}\label{image inpainting}
\min_{f,~ \mathcal{M} \subset \R^{\tau^2}} ~\sum_{x \in \mathscr{I}} \|R_{\mathcal{M},x} (\mathcal{P}(f)) \|_* +\frac{\lambda}{2}\|\nabla_\mathcal{M} f\|_2^2, \quad \text{s.t.} \quad \mathcal{P}(f) \subset \mathcal{M}, f|_\Omega = h|_\Omega.
\end{equation}
In particular, if the index set $\Omega$ is picked as $\{1, s+1, 2s+1,\ldots\} \times \{1, s+1, 2s+1,\ldots\}$, the problem is called sub-sampled super-resolution problem. As the nuclear norm in the first term of the above problem depends on the manifold structure, we consider to solve this problem by alternatively updating the manifold $\mathcal{M}$ and solving $f$ similar as the method considered in \cite{osher2016low}. The outline of solving \eqref{image inpainting} can be stated as follows:
\begin{equation}\label{Inpainting outline}
\begin{cases}
f^{k+1} = \arg\min_{f} \sum_{x \in \mathscr{I}} \|R_{\mathcal{M}^k,x} (\mathcal{P}(f) \|_* +\frac{\lambda}{2}\|\nabla_\mathcal{M}^k f\|_2^2, \\
 \hspace{5cm} \text{s.t.}\quad  \mathcal{P}(f) \subset \mathcal{M}^k, f|_\Omega = h|_\Omega,\\
\mathcal{M}^{k+1} = {P}(f^{k+1}).\\
\end{cases}
\end{equation}
To solve $f^{k+1}$ from the first step in \eqref{Inpainting outline} with a fixed manifold structure $\mathcal{M}^k$, we use the split Bregman iteration~\cite{goldstein2009split}. After introducing an auxiliary variable $\bm{\alpha} = \mathcal{P}(f) \in  \R^{\tau^2 \times mn} $, this problem can be reinterpreted as:
\begin{equation}\label{image inpainting inner ini}
\min_{f,\bm{\alpha}} \sum_{x \in \mathscr{I}} \|R_{\mathcal{M}^k,x} \bm{\alpha} \|_* +\frac{\lambda}{2}\|\nabla_{\mathcal{M}^k} f\|_2^2, ~~ \text{s.t.} ~~ \mathcal{P} (f)  = \bm{\alpha},f|_\Omega = h|_\Omega.
\end{equation}
Since each column of $\bm{\alpha}$ may occur multiply times in different $ \|R_{\mathcal{M}^k,x}\bm{\alpha} \|_*$, it is difficult to simultaneously optimize several nuclear norms together. Therefore, denote the image size as $m \times n$, patch size as $\tau \times \tau$, and the KNN size $K$, we introduce the duplicate operator $\mathcal{Q}: \R^{\tau^2 \times mn}  \rightarrow \R^{K \tau^2 \times mn}$ can be defined as:
 \begin{equation}\label{duplicate operator}
\mathcal{Q} (\bm{\alpha}) = \{ \mathcal{Q}_x (\bm{\alpha}) = R_{\mathcal{M}^k,x} \bm{\alpha}, x \in \mathscr{I} \}
\end{equation}
Then, we denote $\mathcal{Q}_x (\bm{\alpha})= \beta_x, \forall x \in \mathscr{I}$ such that $\|(R_{\mathcal{M}^k,x}) \bm{\alpha} \|_* =  \| {\beta}_x \|_*$. As a result, $\sum_{x \in \mathscr{I}} \|(R_{\mathcal{M}^k,x}) \bm{\alpha} \|_* = \sum_{x \in \mathscr{I}} \|{\beta}_x\|_*$ becomes a separable formula. Thus, Step 1 in \eqref{Inpainting outline} can be reinterpreted as:
\begin{equation}\label{image inpainting inner}
\min_{f,\{\beta_x\}} \sum_{x \in \mathscr{I}} \|{\beta}_x\|_* +\frac{\lambda}{2}\|\nabla_{\mathcal{M}^k} f\|_2^2, ~~ \text{s.t.} ~~\mathcal{Q}_x (\mathcal{P}(f))  = {\beta}_x,f|_\Omega = h|_\Omega.
\end{equation}
Therefore, the above the equality constraint $ \mathcal{Q}_x (\mathcal{P}(f))  = {\beta}_x$ can be solved by considering the following saddle point problem using a augmented Lagrangian formula with the dual variable $\{D_{x}\}$:
\begin{equation}\label{image inpainting inner aug Lag}
\begin{split}
\min_{f,\{\beta_x\}}\max_{\{D_x\}} \sum_{x \in \mathscr{I}} \|{\beta}_x\|_* +\frac{\lambda}{2}\|\nabla_{\mathcal{M}^k} f\|_2^2   + \sum_{x \in \mathscr{I}} \frac{\mu}{2} \|\mathcal{Q}_x (\mathcal{P} (f)) - {\beta}_x + D_{x}\|_2^2, \\
~~ \text{s.t.} ~~  f|_\Omega = h|_\Omega.
\end{split}
\end{equation}
where $\mu$ is the parameter to control the augmented Lagrangian. Similar to the one-step iterative method in the alternating direction method of multipliers (ADMM)
and split Bregman iteration~\cite{glowinski1989augmented, goldstein2009split}, The optimization problem \eqref{image inpainting inner aug Lag} can be iteratively solved as:
\begin{equation}\label{image inpainting inner iteration scheme}
\begin{cases}
\displaystyle \beta_x^{l+1} = \arg\min_{\beta_x}  \|\beta_x\|_* +\frac{\mu}{2}\|{\beta_x} - \mathcal{Q}_x (\mathcal{P}(f^{l})) -D_{x}^{l}\|_2^2, \quad \forall x \in \mathscr{I},\\
\displaystyle f^{l+1} = \arg\min_f \frac{\lambda}{2}\|\nabla_{\mathcal{M}^k} f\|_2^2 +  \sum\limits_{x \in \mathscr{I}} \frac{\mu}{2} \|\mathcal{Q}_x (\mathcal{P} (f)) - {\beta}_x^{l+1}+ D_x^l\|_2^2, ~\text{ s.t.}~  f|_\Omega = h|_\Omega,\\
D_{x}^{l+1} = D_{x}^{l} + (\mathcal{Q}_x(\mathcal{P}(f^{l+1}))-\beta_x^{l+1}), \quad \forall x \in \mathscr{I}.\\
\end{cases}
\end{equation}

The first sub-optimization problem has a closed-form solution provided by the singular value thresholding \cite{cai2010singular}.  Namely,
\begin{equation}\label{SVT for inpainting}
\beta_x^{l+1} = \mathcal{T}_{1/\mu} (\mathcal{Q}_x (\mathcal{P} (f^{l})) +D_x^{l}).
\end{equation}
where for any matrix $X$ with a  singular value decomposition $X = USV$, the singular value thresholding operator $\mathcal{T}$ is provided as
\begin{equation}\label{Singular value threshold}
 \mathcal{T}_{t} (X)=US_{\mathcal{T}}V, \quad S_{\mathcal{T}} = \max(S - t,0).
\end{equation}

Next, we solve $f^{l+1}$ in \eqref{image inpainting inner iteration scheme}, The solution of the linear constrained minimization problem satisfies the following Dirichlet boundary value problem:
\begin{equation}\label{non constraint linear system solving}
\begin{cases}
 \left (- \lambda \Delta_{\mathcal{M}^k} +  \sum\limits_{x \in \mathscr{I}} \mu \mathcal{P}^{\top}\mathcal{Q}_x^{\top}\mathcal{Q}_x \mathcal{P}\right)f  = \mu \mathcal{P}^{\top} \left (  \sum\limits_{x \in \mathscr{I}}  \mathcal{Q}_x^{\top}( {\beta}_x^{l+1} -D_x^{l}) \right ). \vspace{0.2cm}\\
f|_\Omega = h|_\Omega.
\end{cases}
\end{equation}

In \eqref{non constraint linear system solving}, since the duplication operators $\{\mathcal{Q}_x\}$ have only one non-zero element in each row, we have that for all $x$, $(\mathcal{Q}_x^{\top}\mathcal{Q}_x)_{ij} = \sum_{p} (\mathcal{Q}_x^{\top})_{ip} (\mathcal{Q}_x)_{pj}=\sum_{p} (\mathcal{Q}_x)_{pi} (\mathcal{Q}_x)_{pj}$ which is always $0$ if $i \neq j$. Therefore, $\sum\limits_{x \in \mathscr{I}} \mathcal{Q}_x^{\top}\mathcal{Q}_x = W_{\mathcal{Q}}$ becomes a diagonal weight matrix. Similarly, the patch manifold transform operator $\mathcal{P}$ also has only one non-zero element in each row. After left multiplied by a diagonal matrix, $\left( \sum\limits_{x \in \mathscr{I}} \mathcal{Q}_x^{\top}\mathcal{Q}_x \right) \mathcal{P}= W_{\mathcal{Q}}  \mathcal{P}$ is still a matrix with only one non-zero element in each row. Therefore, $\sum\limits_{x \in \mathscr{I}} \mathcal{P}^{\top}\mathcal{Q}_x^{\top}\mathcal{Q}_x \mathcal{P} = W$ is a diagonal weight matrix for the input image whose entries is the occurrence of each pixel in all local regions of patch manifold $\{\beta_x\}$. We can consequently rewrite \eqref{non constraint linear system solving} as:
\begin{equation}\label{linear system solving}
\begin{cases}
 \left (-\lambda \Delta_{\mathcal{M}^k} +  \mu W \right)(f) = \mu \mathcal{P}^{\top} \left (  \sum\limits_{x \in \mathscr{I}}  \mathcal{Q}_x^{\top}( {\beta}_x^{l+1} -D_x^{l}) \right ),\\
f|_\Omega = h|_\Omega,
\end{cases}
\end{equation}
Denote the left hand side of the linear system as $A =  - \lambda \Delta_{\mathcal{M}^k} + \mu W$, plugging the boundary condition $f_\Omega = h_\Omega$ into the first equation, we can solve $f^{l+1}$ restricted in $\Omega^c$ as follows:
\begin{equation}\label{constraint linear system solving}
f^{l+1}|_{\Omega^c} =  (A|_{\Omega^c})^{-1} (\mu \mathcal{P}^{\top} ( \sum\limits_{x \in \mathscr{I}}  \mathcal{Q}_x^{\top}( {\beta}_x^{l+1} -D_x^{l}) - A|_\Omega h|_\Omega).
\end{equation}

Therefore, combining \eqref{Inpainting outline}, \eqref{image inpainting inner iteration scheme}, \eqref{SVT for inpainting} and \eqref{constraint linear system solving}, we can solve the MLR based image inpainting model \eqref{image inpainting} as  Algorithm \ref{alg:Inpainting}. Note that the max number of inner iterations can be chosen as $1$ to reduce the computational time.

\begin{algorithm}

\caption{MLR based image inpainting \eqref{image inpainting}}
\label{alg:Inpainting}
\begin{algorithmic}
\STATE{\textbf{Step 0.}} Using random value to inpaint an initialization of $f^0$ such that $f^0|_\Omega = h|_\Omega$ and corresponding $\mathcal{M}^0$ and $R_{\mathcal{M}^0,x}$ by calculating the KNN of $\mathcal{P}(f^0)$, set $k=0$.
\WHILE{not converge}

\STATE{\textbf{Step 1.0.}} With a fixed $\mathcal{M}^k$, set the initial value of $f^{k+1,0}$ such that $f^{k+1,0}|_\Omega = h|_\Omega$ and calculate the KNN to generate the localize operator $R_{\mathcal{M}^k,x}$, discretized Laplacian operator $\Delta_{\mathcal{M}^k}$, and duplicate operator $\mathcal{Q}$, set $\{\beta_x^{k+1,0}\} = \{\mathcal{Q}_x (\mathcal{P} (f^{k+1,0}))\}$ and $l=0$. Define $A = - \lambda \Delta_{\mathcal{M}^k}  + \mu \sum\limits_{x \in \mathscr{I}} \mathcal{P}^{\top}\mathcal{Q}_x^{\top}\mathcal{Q}_x \mathcal{P}$.
\WHILE{not converge}
\STATE{\textbf{Step 1.1.}} $\beta_x^{k+1,l+1} = \mathcal{T}_{1/\mu} (\mathcal{Q}_x (\mathcal{P} f^{k+1,l}) +D_x^{k+1,l}), \forall x \in \mathscr{I},$
\STATE{\textbf{Step 1.2.}} $f^{k+1,l+1}|_{\Omega^c} =  (A|_{\Omega^c})^{-1} (\mu \mathcal{P}^{\top} ( \sum\limits_{x \in \mathscr{I}}  \mathcal{Q}_x^{\top}( {\beta}_x^{k+1,l+1} -D_x^{k+1,l}) - A|_\Omega h|_\Omega)$,
\STATE{\textbf{Step 1.3.}} $f^{k+1,l+1} = f^{k+1,l+1}|_{\Omega^c} \chi_{\Omega^c} + h \chi_{\Omega},$
\STATE{\textbf{Step 1.4.}} $D_{x}^{k+1,l+1} = D_{x}^{k+1,l} + (\mathcal{Q}_x(\mathcal{P}(f^{k+1,l+1}))-\beta_x^{k+1,l+1}), \forall x \in \mathscr{I},$
\ENDWHILE
\STATE{\textbf{Step 1.5.}} Take $f^{k+1} = f^{k+1,l+1}$,
\STATE{\textbf{Step 2.}} $\mathcal{M}^{k+1} = \mathcal{P}(f^{k+1})$.
\ENDWHILE

\end{algorithmic}
\end{algorithm}

\subsection{MLR for X-ray CT reconstruction}
As a special case of image restoration, medical imaging plays important role in different clinical applications. Here, we consider an application of our method to X-ray Computed Tomography (CT), which aims at reconstructing images from their Radon transform. Mathematically, the X-ray CT reconstruction problem can be essentially represented as a linear inverse problem:
$\mathcal{A}f = g$,
where $\mathcal{A} \in \RR^{m \times n}$ is a measurement matrix representing the collection of discrete line integrations with different projection angles and along different beamlets, $f \in \RR^n$ is vectorized 2 dimensional image and $g \in \RR^m$ is the corresponding measurement. Given the geometry matrix $\mathcal{A}$ and $g$, the task of X-ray CT reconstruction is to find an appropriate value of $f$ \cite{radon1917bestimmung,hsieh2009computed}. In literature, there are some classical methods available, such as the filtered back projection (FBP) type methods \cite{feldkamp1984practical, Defrise1994, Noo1996, Li2006}, the algebraic reconstruction techniques (ART) \cite{gordon1970algebraic}. In practice, however, to minimize the radiation dose by reducing the number of projection angles and beamlets, the amount of measurement $m$ becomes much less than the dimension of the object image $n$, which makes the reconstruction becoming an under-determined problem with infinitely many solutions. As a result, previously mentioned FBP and ART methods usually suffer from artifacts because of the insufficient measurements. Regularization methods such as TV based medical imaging models \cite{TVtomo} and wavelet regularization based medical imaging models \cite{JDLJ2010, DongLiShen2012} makes it possible to reconstruct piecewise smooth or piecewise constant object images. However, it is still a big challenge to preserve tiny features due to possible over-smoothing, which motivate us to propose a MLR CT imaging model to preserve both smooth pieces and tiny features. This model is a special case of \eqref{non local gradient} with linear degradation operator $\mathcal{D} = \mathcal{A}$ and $\lambda = 0$ as follows:
\begin{equation}\label{non local gradient linear}
\min_{f, \mathcal{M} \subset \R^{\tau^2}} ~\sum_{x \in \mathscr{I}}  \| R_{\mathcal{M},x} (\mathcal{P}(f)) \|_* , \quad \text{s.t.}\quad \mathcal{P}(f) \subset \mathcal{M}, \quad \mathcal{A}f = g.
\end{equation}
Note that this model is also applicable for average filter based super resolution problem, Fourier domain inpainting problem, and image deconvolution problems.

To solve \eqref{non local gradient linear}, similar as \eqref{image inpainting inner aug Lag}, after defining the duplication operator $\{\mathcal{Q}_x\}$ and localized patch manifold $\{\beta_x\}$, by splitting the linear constraints $\mathcal{Q}_x(\mathcal{P}(f)) = \beta_x, \forall x \in \mathscr{I}$ and $\mathcal{A}f = g$, we obtain  the saddle point problem of model \eqref{non local gradient linear} using the augmented Lagrangian:
\begin{equation}\label{image  Super Resolution inner Empi aug Lag}
\begin{split}
\min_{f,\bm{\beta}}\max_{\{D_{1,x}\},D_2} \sum_{x \in \mathscr{I}} \|{\beta}_x\|_* + \frac{\mu_1}{2} \sum_{x \in \mathscr{I}} \|\mathcal{Q}_x (\mathcal{P} (f)) - {\beta}_x + D_{1,x}\|_2^2 + \frac{\mu_2}{2} \|\mathcal{A} f    -  g +D_2 \|_2^2.
\end{split}
\end{equation}
Similar as Algorithm \ref{alg:Inpainting}, applying the ADMM we can design algorithm~\ref{alg:SuperResolution} for solving CT reconstruction model \eqref{non local gradient linear}.

\begin{algorithm}

\caption{MLR based CT imaging \eqref{non local gradient linear} }
\label{alg:SuperResolution}
\begin{algorithmic}

\STATE{\textbf{Step 0.}} Using random value to inpaint an initialization of $f^0$ and corresponding $\mathcal{M}^0$ and $R_{\mathcal{M}^0,x}$ by calculating the KNN of $\mathcal{P}(f^0)$, set $k=0$.
\WHILE{not converge}

\STATE{\textbf{Step 1.0.}} With a fixed $\mathcal{M}^k$, set the initial value of $f^0$ and calculate the KNN to generate the localize operator $R_{\mathcal{M}^k,x}$, discretized Laplacian operator $\Delta_{\mathcal{M}^k}$, and duplicate operator $\mathcal{Q}$, set $\bm{\beta}^0 = \mathcal{Q} (\mathcal{P} (f^0))$ and $l=0$.
\WHILE{not converge}
\STATE{\textbf{Step 1.1.}}  $\beta_x^{k+1,l+1} = \mathcal{T}_{1/\mu} (\mathcal{Q}_x (\mathcal{P} f^{k+1,l}) +D_{1,x}^{k+1,l}), \forall x \in \mathscr{I},$
\STATE{\textbf{Step 1.2.}} $f^{l+1} = (\mu_1 W + \mu_2 \mathcal{A}^{\top} \mathcal{A})^{-1}(\mu_1 \mathcal{P}^{\top} (\sum_{x \in \mathscr{I}}  {\mathcal{Q}_x^{\top}}({\beta}_x^{l+1} +D_{1,x}^{l}))+\mu_2 \mathcal{A}^{\top} (g - D_2^l)),$
\STATE{\textbf{Step 1.3.}} $D_{1,x}^{k+1,l+1} = D_{1,x}^{k+1,l} + (\mathcal{Q}_x(\mathcal{P}(f^{k+1,l+1}))-\beta_x^{k+1,l+1}), \forall x \in \mathscr{I},$
\STATE{\textbf{Step 1.4.}} $D_2^{l+1} = D_2^{l} + (\mathcal{A} f^{l+1} - g).$
\ENDWHILE
\STATE{\textbf{Step 1.5.}} Take $f^{k+1} = f^{k+1,l+1}$,
\STATE{\textbf{Step 2.}} $\mathcal{M}^{k+1} = \mathcal{P}(f^{k+1})$,
\ENDWHILE

where $W = \sum_{x \in \mathscr{I}} \mathcal{P}^{\top} \mathcal{Q}^{\top} \mathcal{Q} \mathcal{P}$ is a diagonal weight matrix.

\end{algorithmic}
\end{algorithm}

\section{Semi-supervised learning using MLR}
\label{sec:SemiSuperLearning}
As another advantage of the proposed MLR, this idea can be adapted to handle various data processing problem. Here, we propose the extension of this approach to a semi-supervised learning problem. Many other potential applications in data science will be investigated in our future work.

Semi-supervised learning is a learning paradigm aiming at labeling data from a small amount of labeled training data set~\cite{zhu2009introduction}. Mathematically speaking, given a data set $P = \{x_1,x_2,\ldots,x_n\} \subset \mathbb{R}^{d}$, the semi-supervised learning problem is to find a label function $L: P\rightarrow \{0, 1,2,\ldots,l\}$ representing the label index of the each $x_i$ with given prior knowledge of $L$ in a labeled subset set $S \subset P$. The challenge of a semi-supervised learning problem is to estimate an accurate assignment of $L$ based on a vey small portion information $L(S)$. The general idea of semi-supervised learning is to explore the manifold structure of the data based on an assumption that similar unlabeled samples should be assigned the same classification. Based on this, diffusion based models~\cite{zhu2003semi,shi1509harmonic,shi2016weighted} has been considered to tackle this problem. In this section, we would like to formulate a different way of estimating $L$ from highly insufficient labeled samples based on the MLR method.

Similar as notations discussed in \cite{zhu2003semi,shi1509harmonic,shi2016weighted}, to solve the semi-supervised learning problem, we define the cluster functions $\{\bm{\phi}_i(x)\}$ which is partially assigned from the training data $S$.
$$\bm{\phi}_i(x)= \begin{cases}1, \ \ \ L(x) = i. \\0, \ \ \ \text{otherwise}. \end{cases}, x \in S,\quad  i =0, 1,2,\ldots,l.$$
By viewing $\bm{\phi}_i(x)$ a column vector with length $n$, we obtain a cluster matrix $\Phi = (\phi_0,\cdots,\phi_l) \in \R^{n \times (l+1)}$. Therefore, if we can estimate all the components of $\Phi$, or all $\{\phi_i(x)\}$, the value of all unknown $L(x)$ for $x \in P\backslash S$ can be estimated by:
$$L(x)=\arg\max_i \bm{\phi}_{i}(x), \quad \forall x \in P \backslash S.$$

\begin{figure}[htp]
\centering
\begin{tabular}{c@{\hspace{1pt}}c}
\includegraphics[width=0.48\linewidth]{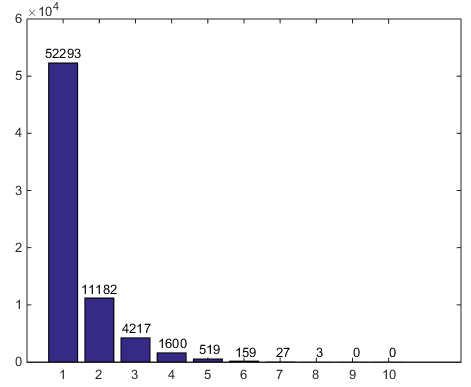} &
\includegraphics[width=0.48\linewidth]{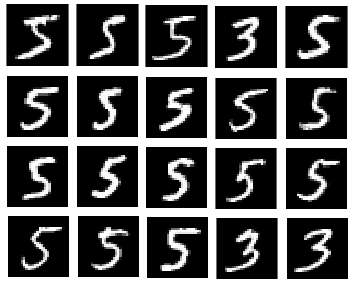} \\
\end{tabular}
\caption{Left Image: The histogram of $\Rank(R_{\mathcal{M},x} \Phi )$ from ground truth of images and labels. Right Image: The $20$-nearest neighborhood of the first point.}
\label{fig:MINSTLocalSimilar}
\end{figure}

Assume the point matrix $P$ is sampled on a manifold $\mathcal{M}$ and define the local restriction operator $R_{\mathcal{M},x}$ as the restriction of a matrix to $x$-th point and its $K$-nearest neighbourhood (KNN). Then by definition of $\Phi$ and $\bm{\phi}_i(x)$, the rank of $R_{\mathcal{M},x} \Phi$ equals to the number of different labels occurred in the KNN. Based on the assumption that similar data samples or nearby points should have similar classification, localization of $\Phi$ should only include a few different labels, i.e., $R_{\mathcal{M},x} \Phi$ has low-rank structure although $\Phi$ might be a full-rank matrix. As an example, we consider the public available MINST data set~\cite{lecun1998gradient} which includes $70,000$ handwritten digits images. We simply view each image as a point in $\RR^d$ and pick the KNNs of each point (image) in terms of Euclidean distance. Left image in Figure \ref{fig:MINSTLocalSimilar} 
shows that majority part of $\{R_{\mathcal{M},x} \Phi \}$ has low-rank structure from the ground truth of cluster matrix $\Phi$. Interestingly, right image in Figure \ref{fig:MINSTLocalSimilar} shows that the $20$-nearest neighborhood of the first image, in which two digits $5$ and $3$ appear because of their similar distribution in terms of Euclidean distance. Therefore, the rank of $R_{\mathcal{M},1} \Phi$ equals to $2$.

Based on the observation that $R_{\mathcal{M},x} \Phi$ has low-rank structure, the corresponding MLR model for cluster matrix estimation can be stated as follows:
\begin{equation}\label{MINST inpainting}
\min_{\Phi} ~\sum_{x \in \mathscr{I}} \|(R_{\mathcal{M},x}) \Phi \|_*, \quad \text{s.t.}\quad  P \subset \mathcal{M}, \quad \Phi(x,i)|_{x \in S} = \begin{cases}1, \ \ \ L(x) = i.\\0, \ \ \ \text{otherwise}. \end{cases}
\end{equation}
Different from the previous image restoration models, the geometric of manifold $\mathcal{M}$ is only determined by information from the data set $P$ which is fixed and irrelevant to the evolution of $\Phi$. Correspondingly, with fixed localization of $\Phi$, the model \eqref{MINST inpainting} is convex and can be solved via standard ADMM. Since it is difficult to simultaneously minimize all the restrictions of $\Phi$, similar as the image restoration cases, we define a duplication operator $\mathcal{Q} = \{\mathcal{Q}_x\}_x$ such that $\mathcal{Q}_x \Phi = R_{\mathcal{M},x}\Phi = \psi_x$ and $\|R_{\mathcal{M},x} \Phi \|_*=\|\psi_x\|_*$. With the auxiliary variables $\{\psi_x\}$ and linear constraint $\mathcal{Q}_x \Phi = \psi_x$, we introduce a group of dual variables $\{D_x\}$ and obtain the following saddle point problem with the augmented Lagrangian:
\begin{equation}\label{MINST inpainting aug Lag}
\begin{split}
\min_{\Phi, \{\psi_x\}}\max_{\{D_x\}} \sum_{x}\left( \|\psi_x\|_*  + \frac{\mu}{2} \|\psi_x - \mathcal{Q}_x (\Phi) -D_{x}\|_2^2 \right) \\ \quad \text{s.t.} \quad   \Phi(x,i)|_{x \in S} = \begin{cases}1, \ \ \ L(x) = i.\\0, \ \ \ \text{otherwise}. \end{cases}
\end{split}
\end{equation}

Similar as the image restoration case, with the definition of the duplication operator $\mathcal{Q}$, because $\mathcal{Q}^{\top}\mathcal{Q} = \sum_{x} \mathcal{Q}_x^{\top}\mathcal{Q}_x = W_{\mathcal{Q}}$ which is a diagonal matrix, we can define the left inverse operator as $\tilde{\mathcal{Q}} = W_{\mathcal{Q}}^{-1} \mathcal{Q}^{\top}$ such that $\tilde{\mathcal{Q}}\mathcal{Q} = I$. Standard ADMM brings the outline of the iteration as follows:
\begin{equation}\label{MINST processing inner iteration scheme}
\begin{cases}
\psi_x^{k+1} = \arg\min_{\psi_x}  \|\psi_x\|_* +\frac{\mu}{2}\|{\psi_x} - \mathcal{Q}_x (\Phi^{k}) -D_{x}^{k}\|_2^2 , \forall x \in P,\\
{\Phi}^{k+1} = \arg\min_\Phi \sum_{x} \frac{\mu}{2}\|{\psi_x^{k+1}} - \mathcal{Q}_x (\Phi) -D_{x}^{k}\|_2^2, ~ \text{s.t.} ~ \Phi(x,i)|_{x \in S} = \begin{cases}1,   \ L(x) = i.\\0, \ \text{otherwise}. \end{cases}\\
D_x^{k+1} = D_x^k + \mathcal{Q}_x (\Phi^{k+1}) - \psi_x^{k+1}, \forall x \in P.
\end{cases}
\end{equation}

In \eqref{MINST processing inner iteration scheme}, the first step can be solved by singular value thresholding operator defined in \eqref{Singular value threshold} as $\psi_x^{k+1} = \mathcal{T}_{1/\mu}(\mathcal{Q}_x (\Phi^{k}) -D_{x}^{k})$. The equality constraint $ \Phi(x,i)|_{x \in S} = \begin{cases}1,  \ \ \ L(x) = i.\\0, \ \ \ \text{otherwise}. \end{cases}$ in the second step is an orthogonal projection operator. Therefore, ${\Phi}^{k+1} = \tilde{\Phi}^{k+1} \chi_{S^c} + \Phi^0 \chi_{S}$, where $\tilde{\Phi}^{k+1} =  (\sum_{x} \mathcal{Q}_x^{\top} \mathcal{Q}_x)^{-1} (\sum_{x} \mathcal{Q}_x^{\top}({\psi_x^{k+1}} - D_{x}^{k})) = W_{{\mathcal{Q}}}^{-1}\sum_{x} \mathcal{Q}_x^{\top}({\psi_x^{k+1}} - D_{x}^{l})) = \tilde{\mathcal{Q}} (\{\psi_x^{k+1}-D_{x}^k\}_x).$ Then the iteration can be re-sketched as:
\begin{equation}\label{MINST processing inner iteration scheme resc}
\begin{cases}
\psi_x^{k+1} = \mathcal{T}_{1/\mu}(\mathcal{Q}_x (\Phi^{k}) -D_{x}^{l}) , \forall x \in P,\\
{\Phi}^{k+1} =  \tilde{\mathcal{Q}} (\{\psi_x^{k+1}-D_{x}^k\}_x) \chi_{S^c} + \Phi^0 \chi_{S},\\
D_x^{k+1} = D_x^k + \mathcal{Q}_x (\Phi^{k+1}) - \psi_x^{k+1}, \forall x \in P.
\end{cases}
\end{equation}

Given an appropriate initialization and sufficient iterations, we obtain the solution of $\Phi$ and the corresponding columns $\phi_i(x)$. Therefore, the index set $L(x)$ for $x \notin S$ can be estimated by $L(x) = \max_{i \in \{0,1,2,\ldots,l\}} \phi_i(x), x \notin S$, which completes the full estimation of $L(x)$.
%
%
%
%
%

It is clear that a better initial guess of $\Phi^0$ can further improve the index completion result. Therefore, we propose to recursively update the initial guess $\Phi^0$ based on the result from \eqref{MINST processing inner iteration scheme resc}, the ultimate algorithm for semi-supervised learning can be summarized in Algorithm \ref{alg:SemiSupervisedLearning}.

\begin{algorithm}

\caption{MLR based semi-supervised learning algorithm}
\label{alg:SemiSupervisedLearning}
\begin{algorithmic}

\STATE{\textbf{Step 0.}} From given point set $P$, generate the manifold $\mathcal{M}$. With a fixed $\mathcal{M}$, calculate the KNN to generate the localize operator $R_{\mathcal{M},x}$. Define the duplicate operator $\mathcal{Q}$ such that $\mathcal{Q} (\Phi) = \Psi = \{(R_{\mathcal{M},x}) \Phi\}$ with $\mathcal{Q}_x (\Phi) = \psi_x = (R_{\mathcal{M},x}) \Phi$. Obtain an initial guess of label function $L^0(x)$ by simply search each unlabeled point's nearest labeled neighborhood and duplicate the label, set $k=0$.
 \WHILE{not converge}
\STATE{\textbf{Step 1.}} Set the initial value of $\Phi^{k+1,0}$ by $\Phi^{k+1,0}(x,i) = \begin{cases}1, \ \ \ L^k(x) = i.\\0, \ \ \ \text{otherwise}. \end{cases}$ , Set the auxiliary variables $\{\psi_x^{k+1,0}\} = 0$ and the dual variable $\{D_{x}^{k+1,0}\} = 0, \forall x$, set $l=0$.
\STATE{\textbf{Step 2.}} Iterating as in \eqref{MINST processing inner iteration scheme resc} to solve a solution of $\Phi^{k+1}$ for model \eqref{MINST inpainting}.
\STATE{\textbf{Step 3.}} Updating $L^{k+1}(x)=\arg\max_i \phi^{k+1}_{i}(x)$.
\ENDWHILE

\end{algorithmic}
\end{algorithm}

\section{Numerical Experiments}
\label{sec:experiments}
In this section, we conduct numerical experiments for the proposed MLR models to various image restoration problems, X-ray CT imaging and semi-supervised learning. Our results validate that the proposed method can successfully reduce the reconstruction error and preserve both edges and repetitive patterns. For all image restoration results, besides the visual quality, we also quantitatively evaluate the results of image restoration using the peak signal-to-noise ratios (PSNR) value:
$$\mbox{PSNR}(f,\tilde{f})=10\log_{10}\frac{MN(f_{\max}-f_{\min})^2}{\|f-\tilde{f}\|_2^2},$$
with the ground truth image $\tilde{f}$, where $f_{\max}$ and $f_{\min}$ are its maximal and minimal pixel values respectively and $M$, $N$ are the size of the image.
All the numerical simulations are implemented by MATLAB in a PC with 32GB RAM and 2.7 GHz CPUs.

\subsection{Image inpainting and super-resolution}

\begin{figure}[thp]
\centering
\begin{tabular}{c@{\hspace{1pt}}c@{\hspace{1pt}}c}
Ground Truth & Incomp. image (5.90 dB) & Harmonic Ext. (22.46 dB)\\
\includegraphics[width=0.31\linewidth]{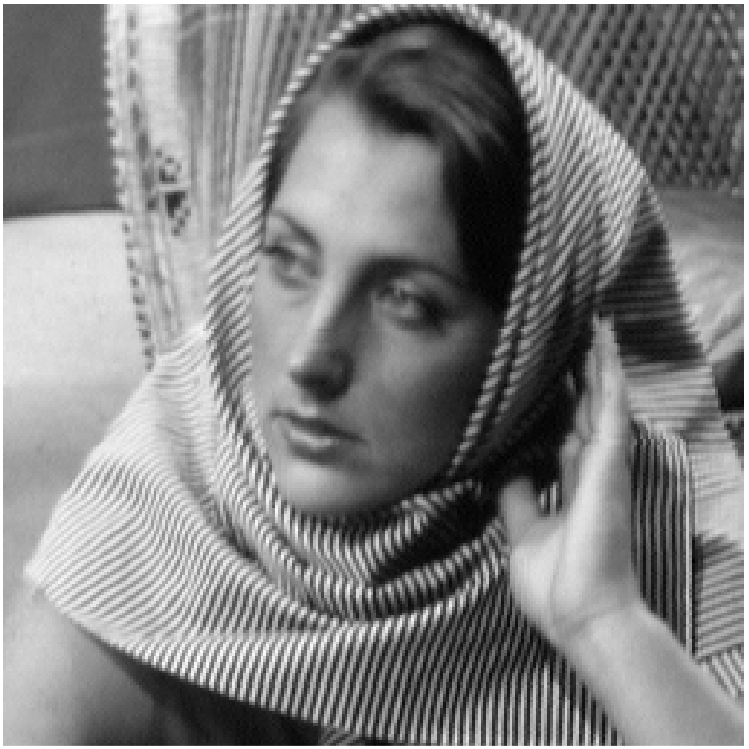} &
\includegraphics[width=0.31\linewidth]{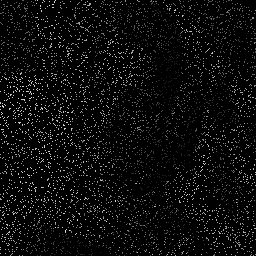} &
\includegraphics[width=0.31\linewidth]{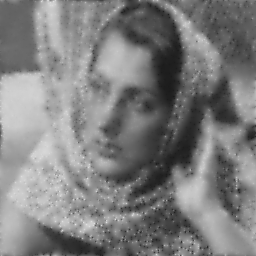} \\
Wavelet (22.83 dB)  & TV (21.97 dB) & MLR ($\lambda = 0$, 22.47 dB) \\
\includegraphics[width=0.31\linewidth]{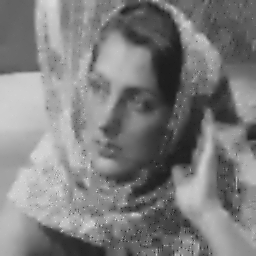} &
\includegraphics[width=0.31\linewidth]{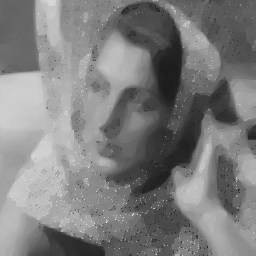} &
\includegraphics[width=0.31\linewidth]{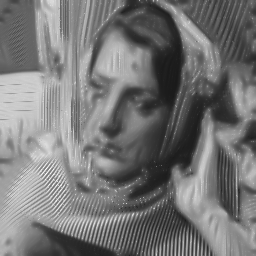} \\
 LDMM (23.73 dB) & LDMM+WGL (25.84 dB) & \textbf{MLR} ($\lambda = -20$, 26.09 dB)\\
\includegraphics[width=0.31\linewidth]{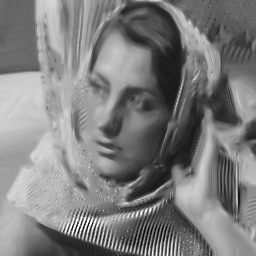} &
\includegraphics[width=0.31\linewidth]{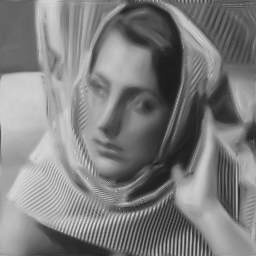} &
\includegraphics[width=0.31\linewidth]{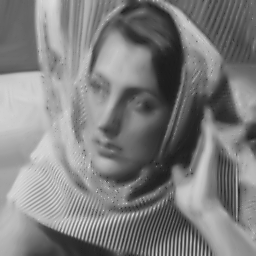}\\
\end{tabular}
\caption{Image inpainting results of $256\times 256$ Barbara image from $10\%$ random available pixels using different methods.}
\label{fig:WaveletDiffusionLowRank}
\end{figure}

In the first experiment, we test Algorithm \ref{alg:Inpainting} to inpaint images from random missing pixels, in which the index set $\Omega$ is uniformly randomly chosen with fixed rate. Figure \ref{fig:WaveletDiffusionLowRank} shows the restoration results of Barbara image from same $10\%$ random available pixels using different methods. It can be seen that the traditional wavelet based method \cite{cai2008framelet}, the classical harmonic extension method and TV based method~\cite{chan2006error} cannot preserve the textures in this low rate of available information because given information in the texture part is recognized as some noise in these two restored images. Both purely manifold based low-rank model and the LDMM method \cite{osher2016low} have much better estimation and preservation of the textures, while the low-rank regularization of the patch manifold may generate some artifacts which breaks some smooth regions. The proposed method include both manifold based low-rank and inverse diffusion ($\lambda = -20$ for image inpainting) can enhance the recovered image to obtain a better texture and smooth region representation. Our method provides comparable results with the most recent proposed LDMM + Weighted graph laplacian (LDMM+WGL) method~\cite{shilow}. 

\begin{figure}[htp]
\centering
\begin{tabular}{c@{\hspace{1pt}}c}
\includegraphics[width=0.45\linewidth]{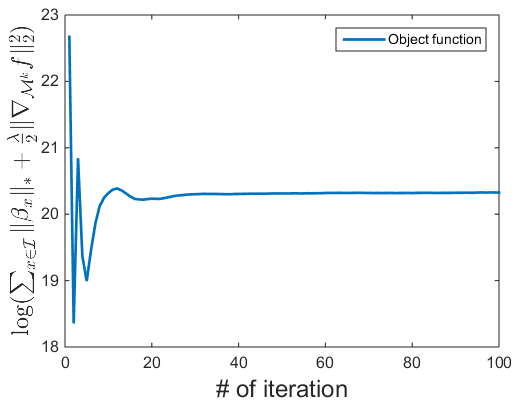}&
\includegraphics[width=0.45\linewidth]{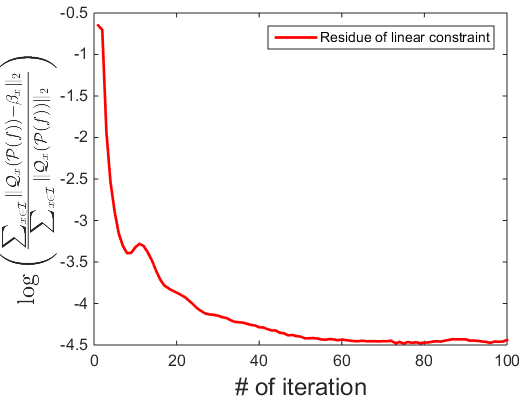}\\
\end{tabular}
\vspace{-0.3cm}
\caption{Convergence curve of Algorithm \ref{alg:Inpainting} for Barbara image inpainting from $10\%$ random sampled pixels. Left: logarithm of object function $\sum_{x \in \mathscr{I}} \|{\beta}_x\|_* +\frac{\lambda}{2}\|\nabla_{\mathcal{M}^k} f\|_2^2$.  Right: logarithm of the relative error between $\{{\mathcal{Q}}_x(\mathcal{P}(f))\}$ and $\{{\beta}_x\}$. }
\vspace{-0.5cm}
\label{fig:convergence curve empirical}
\end{figure}

Due to non-convexity of the model, we also numerically verify the convergence of the algorithm~\ref{alg:Inpainting}. For the numerical simulations shown as above, the convergence curves of the object function $\sum_{x \in \mathscr{I}} \|{\beta}_x\|_* +\frac{\lambda}{2}\|\nabla_{\mathcal{M}^k} f\|_2^2$ and the relative error of linear constraints $ \sum_{x \in \mathscr{I}} \| {\mathcal{Q}}_x( \mathcal{P}(f))- {\beta}_x\|_2$ are shown in Figure \ref{fig:convergence curve empirical}, which validate that for the proposed Algorithm \ref{alg:Inpainting}, the object function converges to a stable value and the relative error of linear constraint converges to zero. 
\begin{figure}[htp]
\vspace{-0.2cm}
\centering
\begin{tabular}{c@{\hspace{1pt}}c@{\hspace{1pt}}c@{\hspace{1pt}}c}
Incomplete image  & LDMM &LDMM+WGL &\textbf{MLR method}\\
\includegraphics[width=0.24\linewidth]{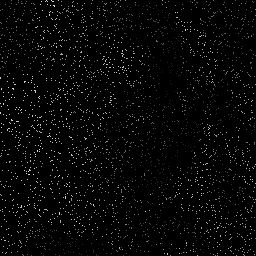}&
\includegraphics[width=0.24\linewidth]{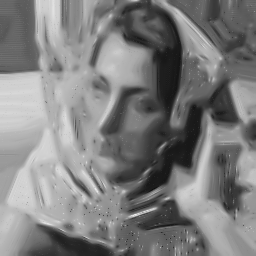} &
\includegraphics[width=0.24\linewidth]{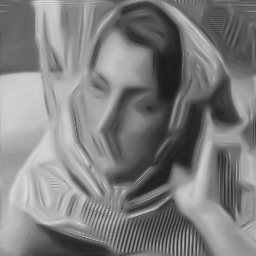} &
\includegraphics[width=0.24\linewidth]{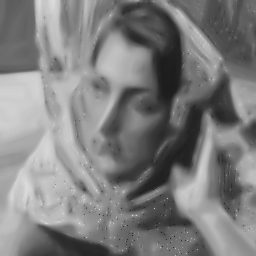}\\
5.66dB& 21.73dB & 23.22 dB &22.17dB\\
\includegraphics[width=0.24\linewidth]{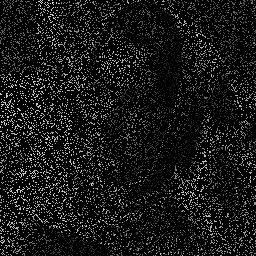}&
\includegraphics[width=0.24\linewidth]{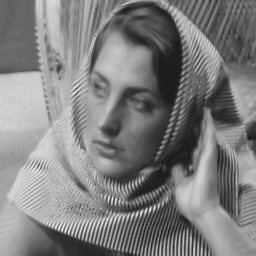} &
\includegraphics[width=0.24\linewidth]{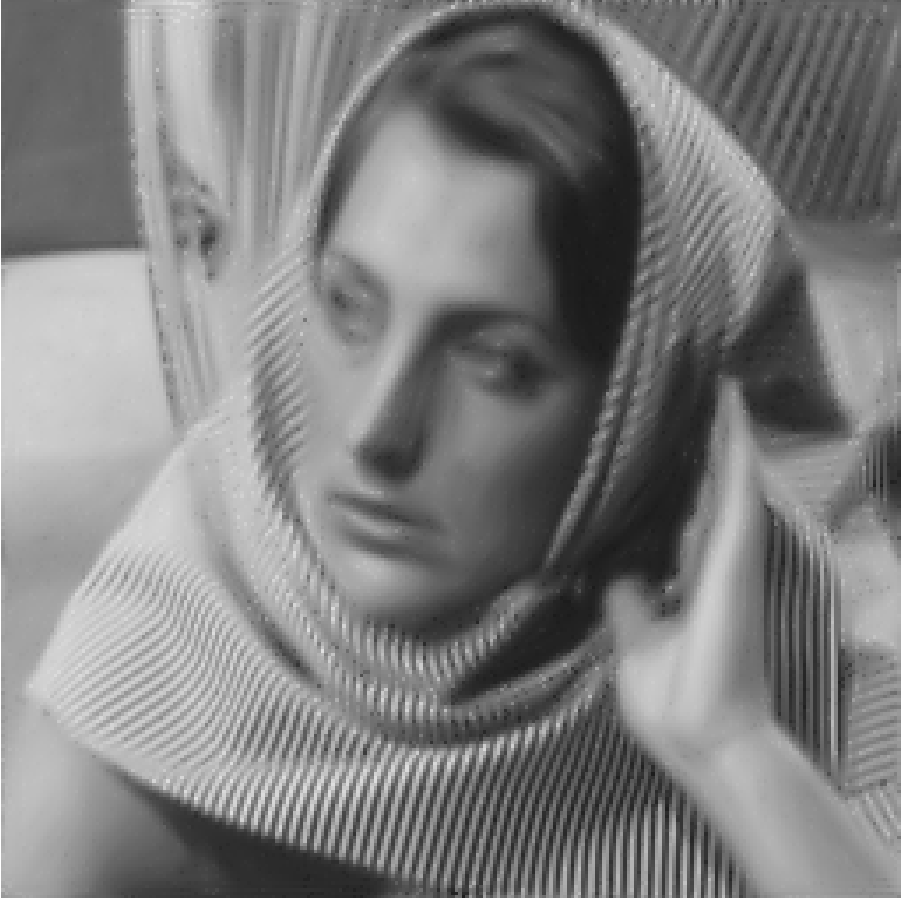} &
\includegraphics[width=0.24\linewidth]{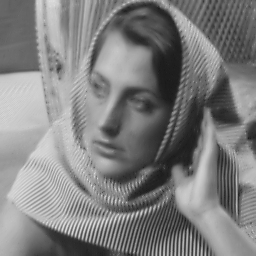}\\
6.41dB&28.11dB&  28.82dB & 29.17dB\\
\includegraphics[width=0.24\linewidth]{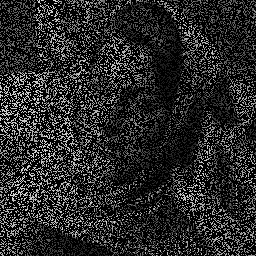}&
\includegraphics[width=0.24\linewidth]{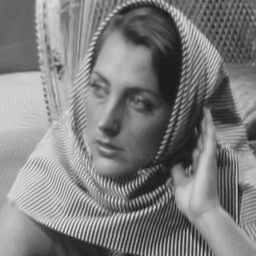} &
\includegraphics[width=0.24\linewidth]{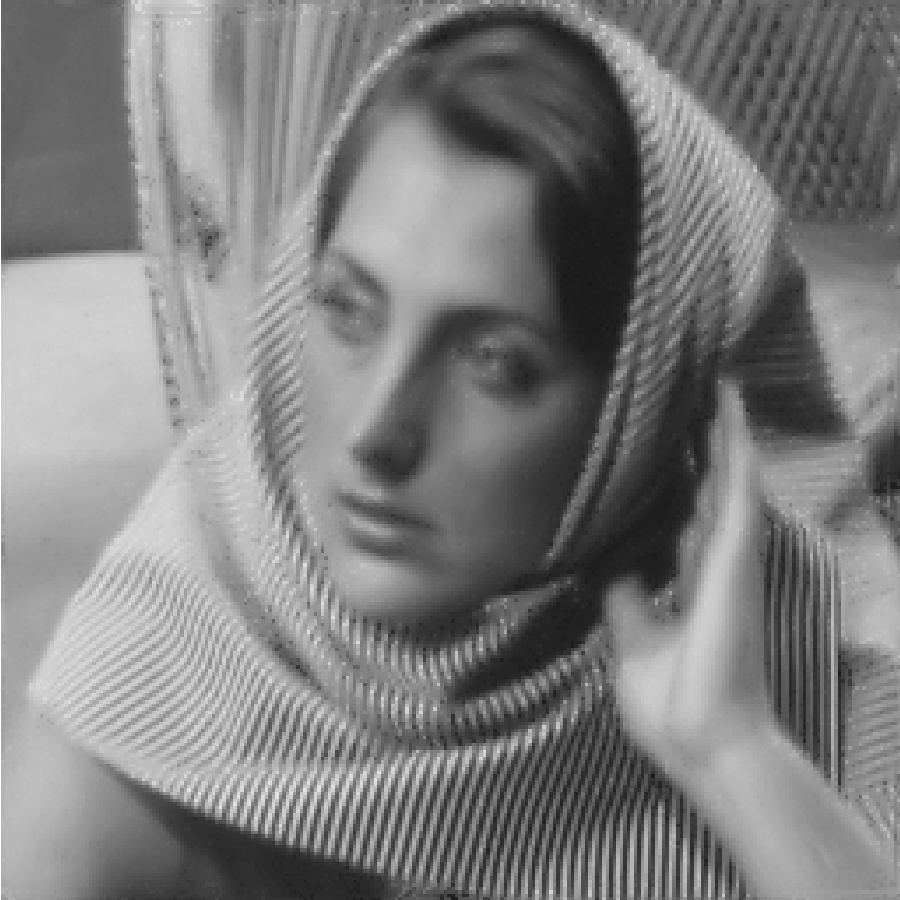} &
\includegraphics[width=0.24\linewidth]{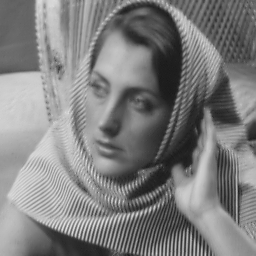}\\
7.16dB&32.02dB& 32.06 dB & 32.05dB\\
\end{tabular}
\vspace{-0.3cm}
\caption{Image inpainting for $256\times 256$ Barbara image. From top to bottom shows the image inpainting results from $5\%, 20\%$ and $40\%$ random available pixels.}
\label{fig:Inpainting Barbara}
\end{figure}

\begin{figure}[htp]
\centering
\begin{tabular}{cc@{\hspace{.5pt}}c@{\hspace{.5pt}}c@{\hspace{.5pt}}c}
& Fingerprint & Boat & Baboon & Peppers\\
\parbox[t]{1mm}{\rotatebox[origin=c]{90}{Ground Truth\hspace{-3cm}}} &
\includegraphics[width=0.23\linewidth]{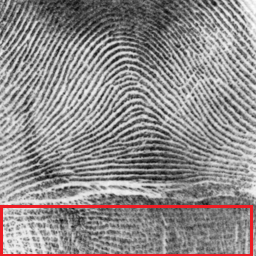}&
\includegraphics[width=0.23\linewidth]{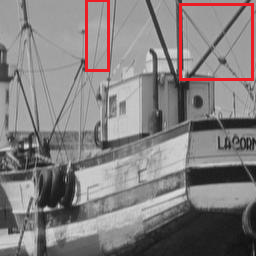}&
\includegraphics[width=0.23\linewidth]{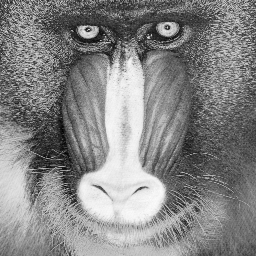}&
\includegraphics[width=0.23\linewidth]{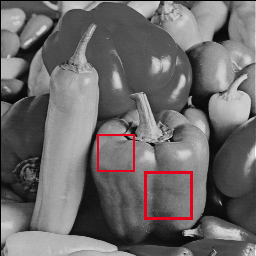}\\
\hline \vspace{-0.3cm}\\
& 5.04dB & 5.70dB & 5.38dB & 6.02dB \\
\parbox[t]{1mm}{\rotatebox[origin=c]{90}{Incomplete image\hspace{-3cm}}} &
\includegraphics[width=0.23\linewidth]{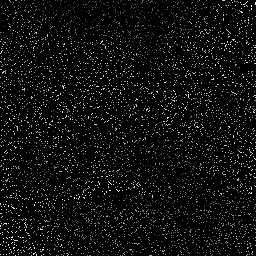}&
\includegraphics[width=0.23\linewidth]{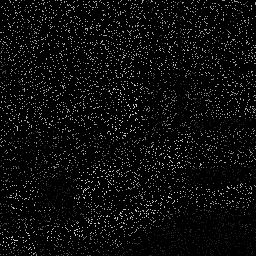}&
\includegraphics[width=0.23\linewidth]{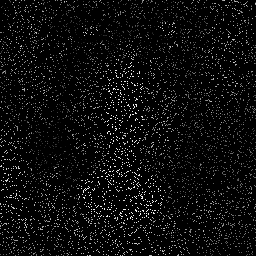}&
\includegraphics[width=0.23\linewidth]{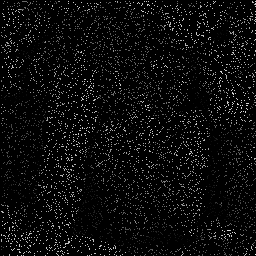}\\
\hline \vspace{-0.3cm}\\
& 19.32dB & 25.08dB &  19.43dB &  23.39dB\\
\parbox[t]{1mm}{\rotatebox[origin=c]{90}{LDMM \hspace{-3cm}}} &
\includegraphics[width=0.23\linewidth]{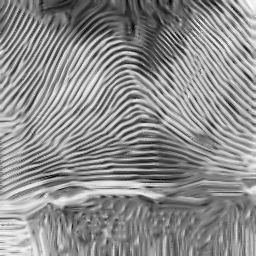} &
\includegraphics[width=0.23\linewidth]{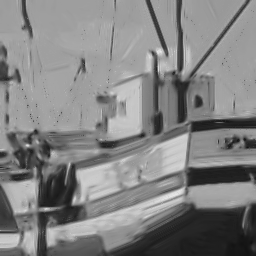} &
\includegraphics[width=0.23\linewidth]{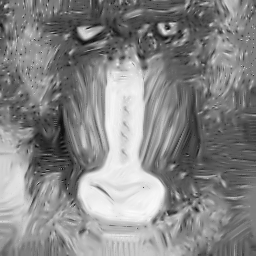} &
\includegraphics[width=0.23\linewidth]{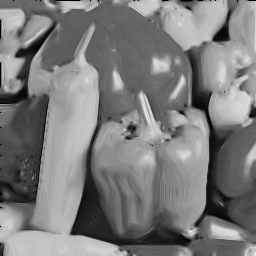} \\
\hline \vspace{-0.3cm}\\
& 20.25 dB & 25.51 dB & 19.79dB & 24.58 dB\\
\parbox[t]{1mm}{\rotatebox[origin=c]{90}{LDMM+WGL\hspace{-3cm}}} &
\includegraphics[width=0.23\linewidth]{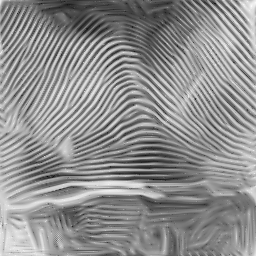} &
\includegraphics[width=0.23\linewidth]{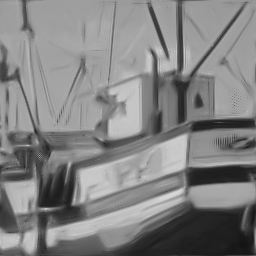} &
\includegraphics[width=0.23\linewidth]{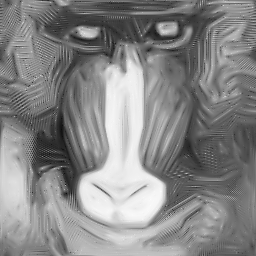} &
\includegraphics[width=0.23\linewidth]{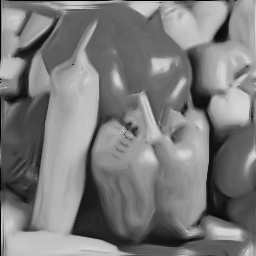} \\
\hline \vspace{-0.3cm}\\
& 20.24dB & 25.77dB & 20.05dB & 24.29dB \\
\parbox[t]{1mm}{\rotatebox[origin=c]{90}{\textbf{MRL}\hspace{-3cm}}} &
\includegraphics[width=0.23\linewidth]{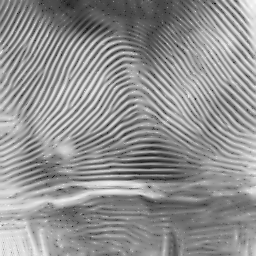} &
\includegraphics[width=0.23\linewidth]{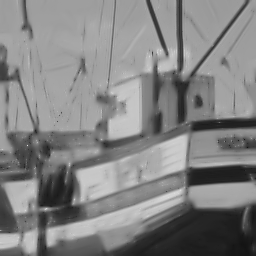} &
\includegraphics[width=0.23\linewidth]{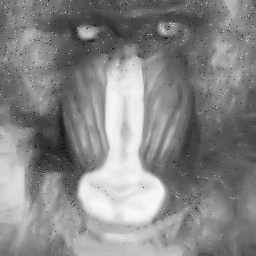} &
\includegraphics[width=0.23\linewidth]{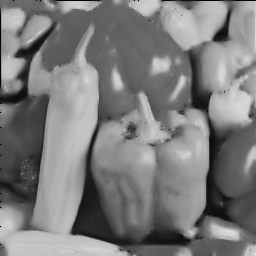} \\
\end{tabular}
\caption{Image inpainting for different images from $10\%$ available pixels. From top to bottom: Ground truth, incomplete images, results from LDMM~\cite{osher2016low}, results from LDMM+WGL~\cite{shilow}, results from MRL, respectively.}
\label{fig:InpaintingCompTextureCartoon}
\end{figure}

We further test the proposed model for different level of available information and conduct comparisons with the LDMM method. Figure \ref{fig:Inpainting Barbara} shows other Barbara image inpainting results from $5\%, 20\%$ and $40\%$ random available information. In the case of using $5\%$ available information, the MLR model produces a qualitatively and quantitatively better result than the one obtained from LDMM. However, the image from LDMM+WGL method has the highest PSNR, although it qualitatively produces more artifacts near the mouth region. In the case of using $10\%$ available information, although the proposed MLR model produces an image with the highest PSNR value, it is hard visually distinct results from MLR and LDMM+WGL. Thus, MLR and LDMM+WGL are comparable and better than LDMM in this case. MLR and LDMM+WGL methods produce similar high quality results when the sampling rate increases to $20\%$ available information although this rate of information may also be quite challenging to other existing methods. All three methods produces very good results with $40\%$ information. Moreover, we also apply the proposed image inpainting model to other images to test the capability of the MLR for handling texture and carton parts. For images with more textures such as the fingerprint image, the baboon image and the boat image, Figure \ref{fig:InpaintingCompTextureCartoon} shows that the proposed MLR method can still preserve more features. In particular, at the bottom part of the fingerprint image highlighted by the red box, the LDMM method generates some vertical artifacts while the MLR method produce more accurate estimation. The LDMM+WGL method successfully improves the inpainting results from the LDMM method, but some vertical artifacts still remain. For the boat image in Figure \ref{fig:InpaintingCompTextureCartoon}, we observe that the proposed MLR method can restore more isolated line structures on the top of the boat as highlighted by red boxes while the LDMM method tends to remove the thin lines. The LDMM+WGL method produces a comparable result with the one from MLR method. For the baboon image, since the texture is too tiny and not repeated frequently, all methods do not provide a result with clear skin and beard structure. The LDMM+WGL method seems to enlarge the artifacts in this case. On the other hand, for images with less textures such as the peppers image, Figure \ref{fig:InpaintingCompTextureCartoon} shows that the proposed method can reduce the possibility of generating artifacts which should not exist. For example, at the center of the green pepper (highlighted by the red box), and at the center of the camera support (highlighted by the red box), the artifacts from the LDMM method and the LDMM+WGL method break the smooth regions while the proposed MLR method preserves the smooth parts because the smooth regions also include repetitive patterns and formulate the low-rank structure.

\begin{figure}[htp]
\begin{center}
\begin{minipage}{0.32\linewidth}
\centering {Incomplete image}\\
\includegraphics[width=1\linewidth]{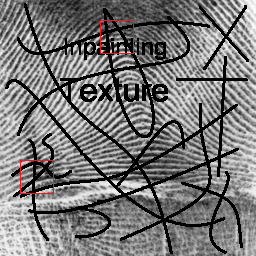}
\end{minipage}
\begin{minipage}{0.32\linewidth}
\centering {Wavelet model~\cite{cai2008framelet}}\\
\includegraphics[width=1\linewidth]{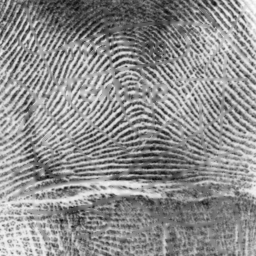}
\end{minipage}
\begin{minipage}{0.32\linewidth}
\centering {\textbf{MLR method}}\\
\includegraphics[width=1\linewidth]{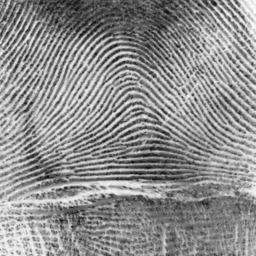}
\end{minipage}
\begin{minipage}{0.32\linewidth}
\includegraphics[width=0.48\linewidth]{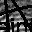}
\includegraphics[width=0.48\linewidth]{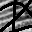}\\
\centering{12.73dB}
\end{minipage}
\begin{minipage}{0.32\linewidth}
\includegraphics[width=0.48\linewidth]{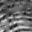}
\includegraphics[width=0.48\linewidth]{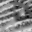}\\
\centering{24.91dB}
\end{minipage}
\begin{minipage}{0.32\linewidth}
\includegraphics[width=0.48\linewidth]{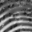}
\includegraphics[width=0.48\linewidth]{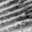}\\
\centering{29.97dB}
\end{minipage}\\
\begin{minipage}{0.32\linewidth}
\includegraphics[width=1\linewidth]{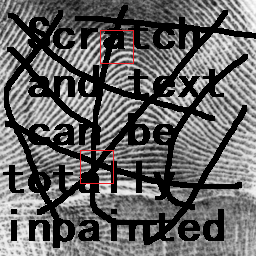}
\end{minipage}
\begin{minipage}{0.32\linewidth}
\includegraphics[width=1\linewidth]{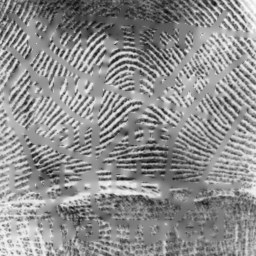}
\end{minipage}
\begin{minipage}{0.32\linewidth}
\includegraphics[width=1\linewidth]{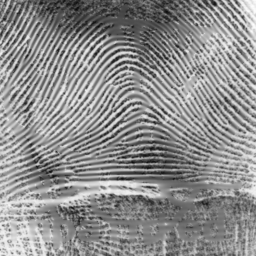}
\end{minipage}
\begin{minipage}{0.32\linewidth}
\includegraphics[width=0.48\linewidth]{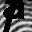}
\includegraphics[width=0.48\linewidth]{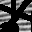}\\
\centering{9.97dB}
\end{minipage}
\begin{minipage}{0.32\linewidth}
\includegraphics[width=0.48\linewidth]{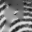}
\includegraphics[width=0.48\linewidth]{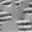}\\
\centering{21.24dB}
\end{minipage}
\begin{minipage}{0.32\linewidth}
\includegraphics[width=0.48\linewidth]{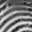}
\includegraphics[width=0.48\linewidth]{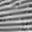}\\
\centering{25.74dB}
\end{minipage}\\
\end{center}
\caption{Image inpainting for Fingerprint image corrupted by two different type of scratches and texts. The bottom images of each row include two zoom-in regions highlighted by red boxes.}
\label{fig:Inpainting Scratch}
\end{figure}

Additionally, we also implement the MLR method for image inpainting from manual scratches. Figure \ref{fig:Inpainting Scratch} shows that compared to the wavelet based image inpainting model \cite{cai2008framelet}, the proposed model has much better quality of recovering the fingerprint structure in terms of both the visualization and the PSNR value. Moreover, for the second row with wider scratches, the proposed MLR model has better estimation of the fingerprint pattern other than simply smoothen the scratched regions.

\begin{figure}[h]
\begin{center}
\begin{tabular}{c@{\hspace{1pt}}c@{\hspace{1pt}}c@{\hspace{1pt}}c}
Bi-Cubic interp. & LDMM & LDMM+WGL &\textbf{MLR method}\\
\includegraphics[width=0.24\linewidth]{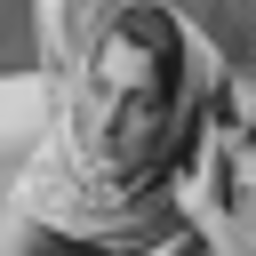}&
\includegraphics[width=0.24\linewidth]{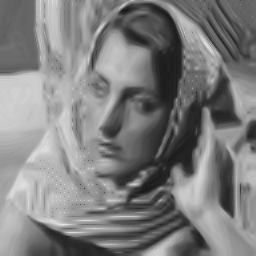} &
\includegraphics[width=0.24\linewidth]{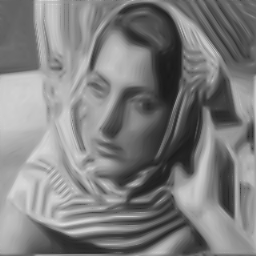} &
\includegraphics[width=0.24\linewidth]{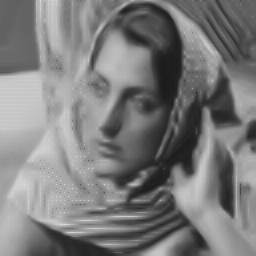}\\
21.06dB & 21.64dB &  21.32dB &  21.87dB\\
\includegraphics[width=0.24\linewidth]{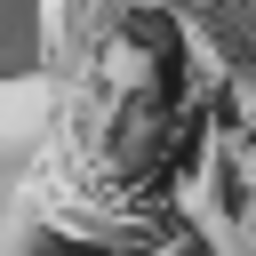}&
\includegraphics[width=0.24\linewidth]{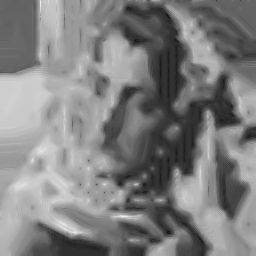} &
\includegraphics[width=0.24\linewidth]{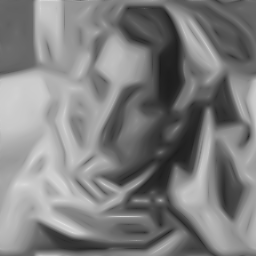} &
\includegraphics[width=0.24\linewidth]{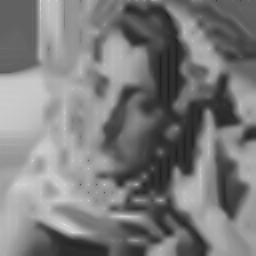}\\
19.08dB & 20.21dB & 20.31dB & 20.47dB\\
\end{tabular}
\end{center}
\caption{Super resolution from image subsampling.  From top to bottom shows the image super resolution results for down sample rate $4 \times 4$ and $8 \times 8$. From left to right shows the result from Bi-cubic interpolation, low-dimension manifold based method and the proposed MLR method.}
\label{fig:SuperResolution}
\end{figure}

In the second experiment, we show the results of super-resolution. In \cite{osher2016low}, the authors conduct the super resolution as a special type of image inpainting problem with highly coherent fixed index set $\Omega=\{1,s+1,2s+1,\ldots\} \times \{1,s+1,2s+1,\ldots\}$. Using the same model and algorithm as the image inpainting problem, the results of this super-resolution problem from sub-sampled pixel are shown as follows in \ref{fig:SuperResolution}. It can be seen that the super-resolution result is better than results from traditional bi-cubic interpolation and comparable to results from the LDMM method and the LDMM+WGL method. 

\begin{figure}[htp]
\centering
\begin{tabular}{c@{\hspace{1pt}}c@{\hspace{1pt}}c}
Bi-Cubic interpolation & LDMM method  &\textbf{MLR method}\\
\includegraphics[width=0.32\linewidth]{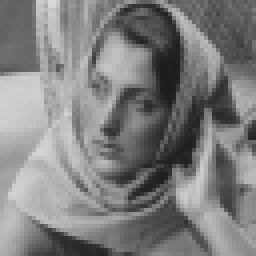}&
\includegraphics[width=0.32\linewidth]{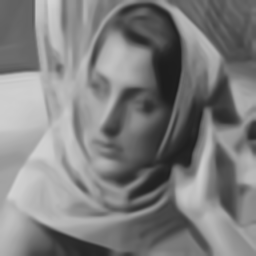} &
\includegraphics[width=0.32\linewidth]{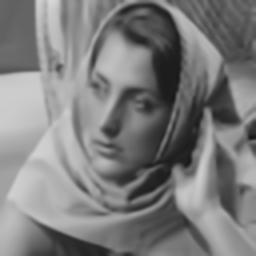}\\
22.93dB & 23.52dB & 23.71dB\\
\includegraphics[width=0.32\linewidth]{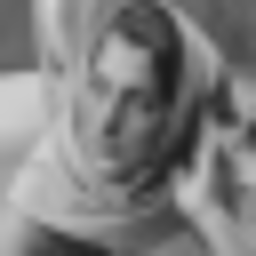}&
\includegraphics[width=0.32\linewidth]{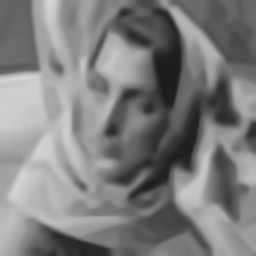} &
\includegraphics[width=0.32\linewidth]{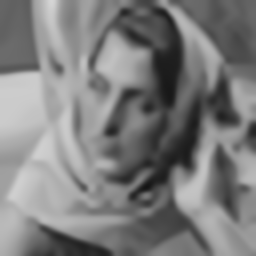}\\
21.61dB & 22.33dB & 22.42dB\\
\end{tabular}
\caption{Super resolution from average. From top to bottom shows the image super resolution results for down sample rate $4 \times 4$ and $8 \times 8$. From left to right shows the result from Bi-cubic interpolation, LDMM method and the proposed MLR method.}
\label{fig:SuperResolutionAV}
\end{figure}

As another case of super resolution, the problem is assumed as image restoration from filtered low resolution version of images. Define an average operator $\mathcal{A}$, the input low resolution image $f_L = \mathcal{A}(f)$, which provide a linear constraint fidelity condition and similar as the medical imaging model \eqref{non local gradient linear}. Using the formula \eqref{image  Super Resolution inner Empi aug Lag} and applying Algorithm \ref{alg:SuperResolution}, the super resolution results from $4 \times 4$ and $8 \times 8$ average filtered low resolution images are shown in Figure \ref{fig:SuperResolutionAV}. The proposed MLR method produces more detailed information and sharper images than bi-cubic interpolation and LDMM in \cite{osher2016low}. 

\subsection{X-ray CT Reconstruction}
\begin{figure}[htp]
\centering
\includegraphics[width=0.31\linewidth]{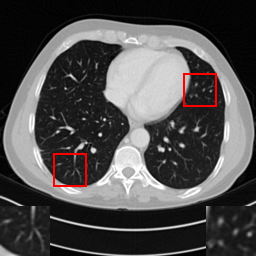}\\
\begin{tabular}{c@{\hspace{1pt}}c@{\hspace{1pt}}c}
15 Projections & 30 Projections & 60 Projections\\
\includegraphics[width=0.32\linewidth]{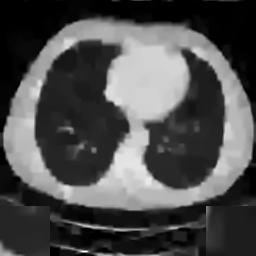}&
\includegraphics[width=0.32\linewidth]{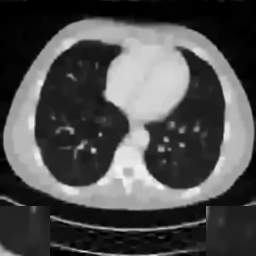}&
\includegraphics[width=0.32\linewidth]{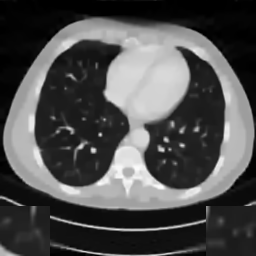}\\
20.83dB & 23.79dB &25.81dB\\
\includegraphics[width=0.32\linewidth]{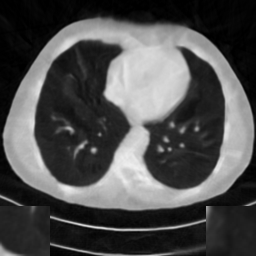}&
\includegraphics[width=0.32\linewidth]{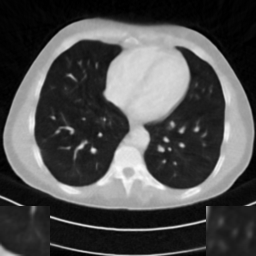}&
\includegraphics[width=0.32\linewidth]{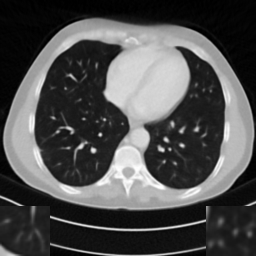}\\
24.04dB&28.08dB&31.29dB\\
\end{tabular}
\caption{Fan-beam imaging for a clinical X-ray scanned chest slice from 15, 30 and 60 projection views. The first row shows the ground truth image from a resized ($256\times 256$) human chest slice, where the left bottom and right bottom images are the zoom-in views of the regions enclosed by the red rectangles. The second row shows the results obtained from the wavelet tight frame based model~\cite{DongLiShen2012}. The third row shows the results obtained from the proposed MLR based method. }
\label{fig:Fan beam Phantom}
\end{figure}

It is quite challenging to reconstruct satisfactory image for the X-ray CT problem with a small amount of radiation dose.
In this section, we apply the model \eqref{image Super Resolution inner Empi aug Lag} and Algorithm \ref{alg:SuperResolution} to the fan-beam projection measurement of images with reduced number of projection views. We consider the CT imaging for a human chest slice (See Figure \ref{fig:Fan beam Phantom})  from the data of "Low Dose CT Grand Challenge" provided by Dr. Cynthia McCollough, the Mayo Clinic, the American Association of Physicists in Medicine, and supported by grants EB017095 and EB017185 from the National Institute of Biomedical Imaging and Bioengineering. Regarding to the linear fidelity $\mathcal{A}f=g$, the ground truth image and the object image $f$ has resolution $256 \times 256$ and the Radon transform measurement $g$ in this section always includes $512$ projection lines in each projection view. Therefore, $\#PROJ$ projection views represents the measurements has $\frac{\text{Card}(g)}{\text{Card}(f)}=\frac{\#PROJ\times 512}{256^2} =\frac{\#PROJ}{128} $ portion of the object function. The huge sparse geometric matrix $\mathcal{A}$ is generated by Siddon's method~\cite{Siddon} as pre-process.

For CT imaging from $15$, $30$ and $60$ views, the CT reconstruction results from the proposed MLR method are shown in Figure \ref{fig:Fan beam Phantom}. It can be seen that the proposed model performs better than the wavelet based method~\cite{DongLiShen2012} in term of both the visual quality and the PSNR value. For wavelet based method, stronger regularization as in Figure \ref{fig:Fan beam Phantom} would remove the small features since they would be recognized as artifacts or noise, while weaker regularization cannot remove the artifacts caused by insufficient projection angles. In particular, in the case with 15 projections, the wavelet based method cannot recover the main vessels at the right side while our method still produce very good results. Moreover, for 60 projections, the zoom-in part shows that the proposed model can successfully reconstruct these tiny features, which is important for futher clinical diagnosis and therapy.

\begin{figure}
\centering
\begin{tabular}{c@{\hspace{1pt}}c@{\hspace{1pt}}c}
15 Projections & 20 Projections & 30 Projections\\
\includegraphics[width=0.32\linewidth]{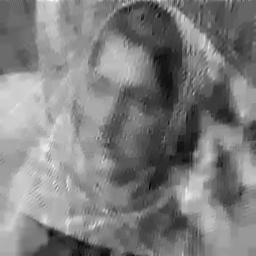}&
\includegraphics[width=0.32\linewidth]{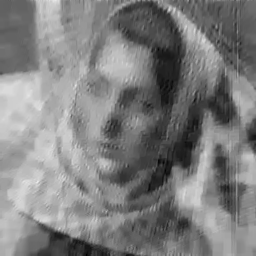}&
\includegraphics[width=0.32\linewidth]{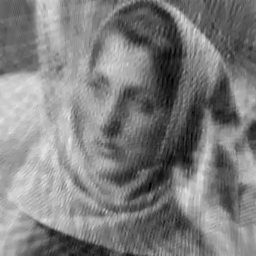}\\
21.69dB  & 22.39dB  &23.37dB\\
\includegraphics[width=0.32\linewidth]{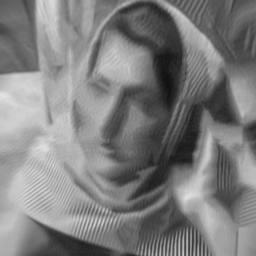}&
\includegraphics[width=0.32\linewidth]{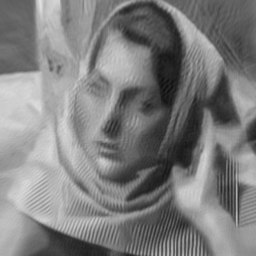}&
\includegraphics[width=0.32\linewidth]{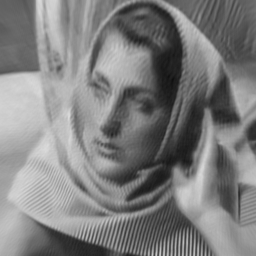}\\
23.25dB&24.50dB&25.84dB\\
\end{tabular}
\caption{Fan-beam imaging for Barbara image from 15, 20 and 30 projection views, respectively. The first row shows the result by wavelet tight frame based model. The second row shows the result from the proposed MLR based method. }
\label{fig:Fan beam Barbara1}
\end{figure}

Additionally, to further illustrate the effectiveness of the MLR method for CT image reconstruction, we test our method by applying the geometric matrix to a natural image. Figure \ref{fig:Fan beam Barbara1} show that for inverse Radon transform of natural image with apparent textures from all $15$, $20$ and $30$ projection views, the proposed method has even greater advantage comparing to the wavelet based CT reconstruction method since the traditional wavelet based method cannot distinguish the texture from the artifacts caused by the low-dose projection.

\subsection{Semi-supervised Learning}
\begin{figure}[h]
\centering
\includegraphics[width=0.95\linewidth]{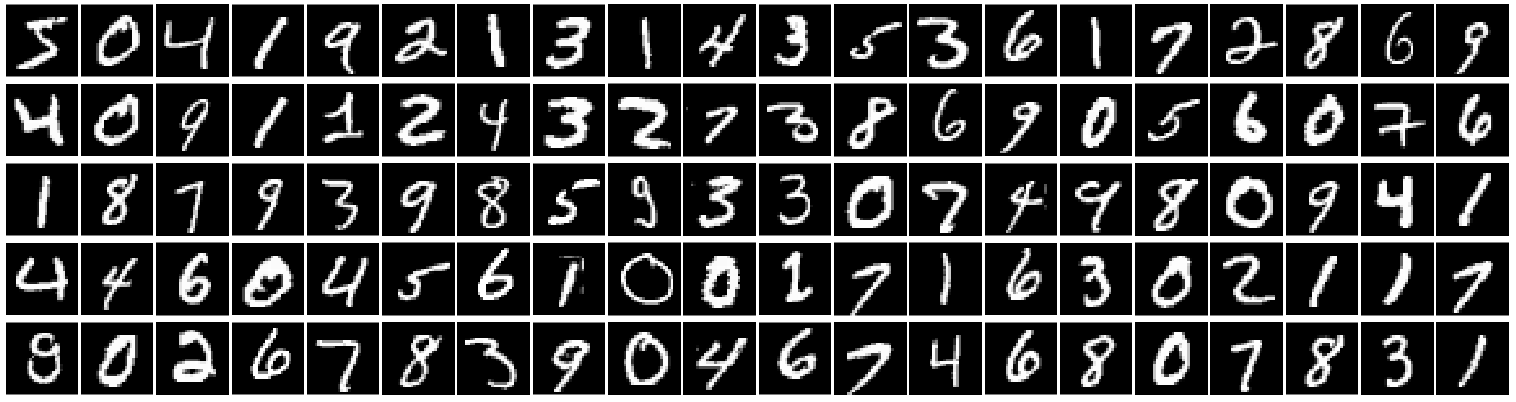}\\
\caption{First 100 "hand writing number" images of MINST data.}
\label{fig:MINST}
\end{figure}

Our final experiment is conducted to test the proposed MLR method for handwritten digits recognition based on the MINST data which is initially provided and processed in \cite{lecun1998gradient}, as shown in Figure \ref{fig:MINST}, including totally 70, 000 different $28 \times 28$ ``handwritten digits" images. As a special case of semi-supervised learning problem, we regard each image as a $784$ dimensional vector, and view all the images as a set of 70, 000 points in $\R^{784}$. Therefore, the vectorized images can formulate a point matrix $P \in \R^{784 \times 70000}$. The labels $\{L(x)\}$ can possibly take the value from ${0,1,2,\ldots,9}$.

For initial purpose of MINST data, the given indices set $S$ has size $60,000$ and one need to estimate the rest $10,000$ index with lowest error. Recently, the full $70,000$ indices set can be roughly reconstructed from $50 - 100$ given indices and some diffusion based methods. For example, \cite{zhu2003semi} proposed an initial graph Laplacian based method. Later on, \cite{shi2016weighted} proposed a weighted graph Laplacian method, from which the inpainting accuracy can exceed $80\%$ from merely $70$ of given indices.

\begin{figure}[htp]
\centering
\begin{tabular}{c@{\hspace{1pt}}c@{\hspace{1pt}}c}
35 Samples& 50 Samples & 70 Samples\\
\includegraphics[width=0.32\linewidth]{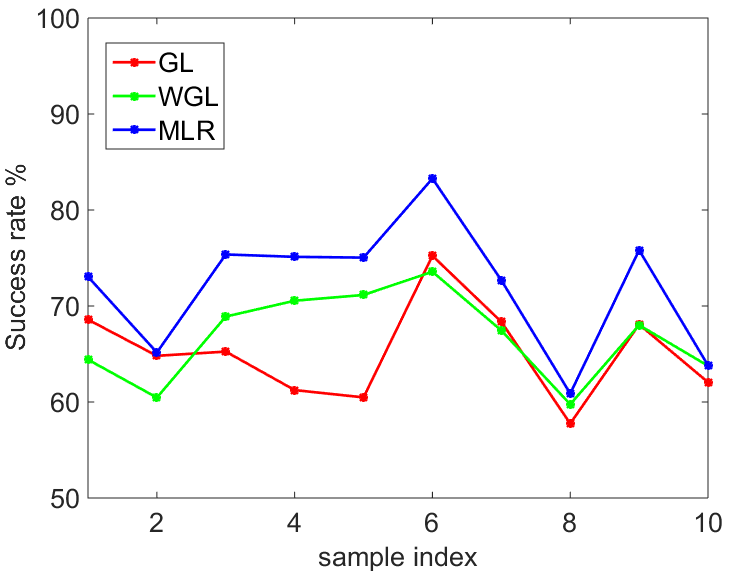} &
\includegraphics[width=0.32\linewidth]{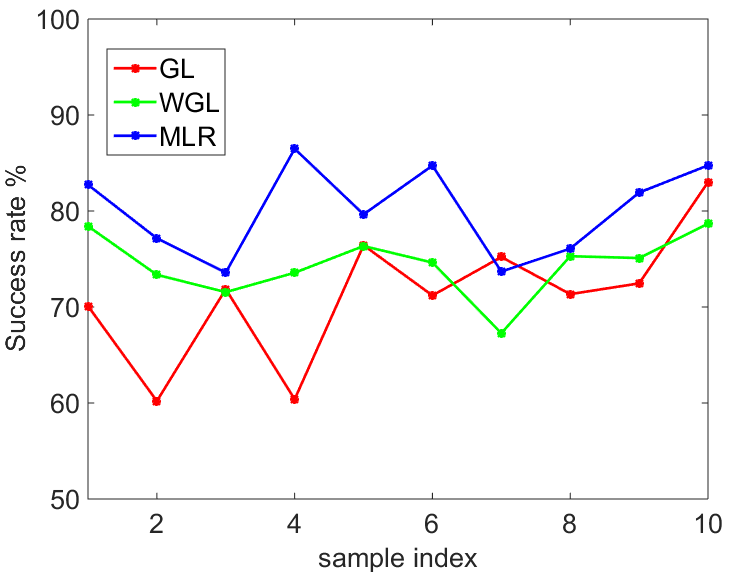} &
\includegraphics[width=0.32\linewidth]{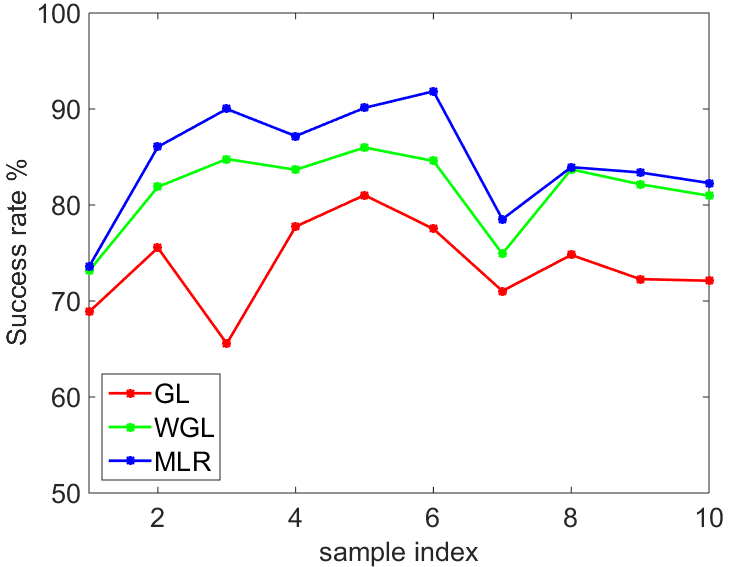}\\
100 Samples& 700 Samples & Average\\
\includegraphics[width=0.32\linewidth]{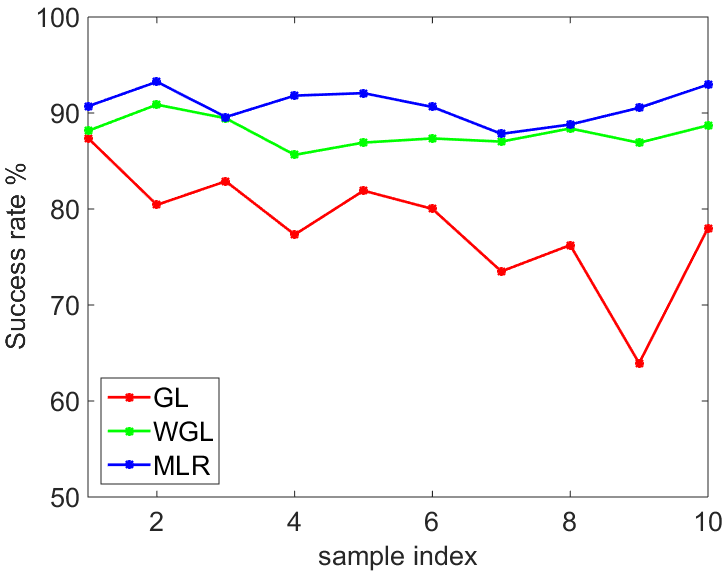} &
\includegraphics[width=0.32\linewidth]{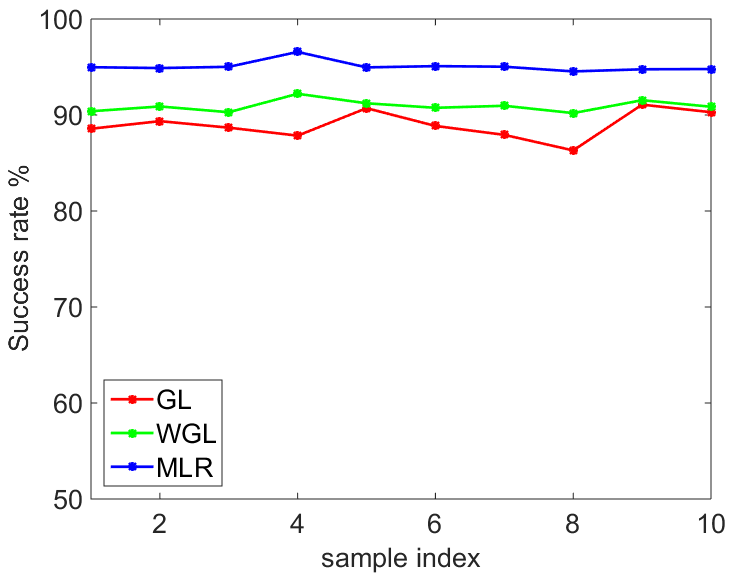} &
\includegraphics[width=0.32\linewidth]{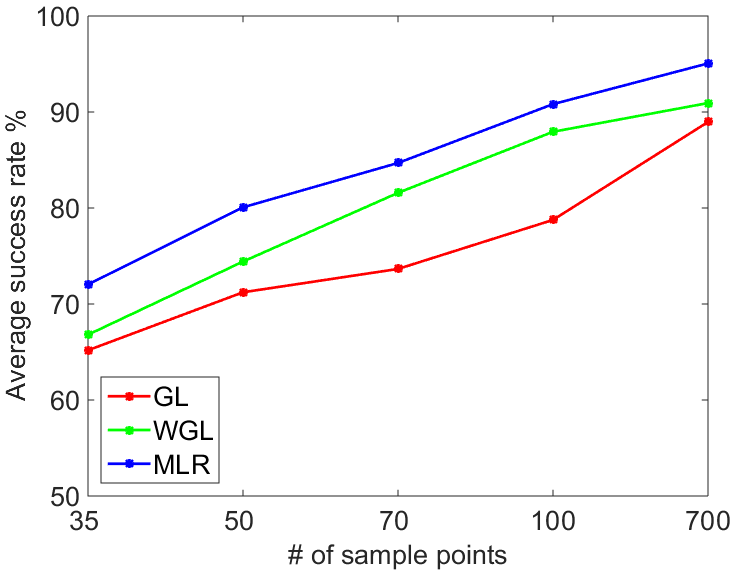}\\
\end{tabular}
\caption{Success rate of label estimation by graph Laplacian, weighted graph Laplacian, and proposed MLR methods.}
\label{fig:MINSTResult}
\end{figure}

In this experiment, we apply the MLR based Algorithm \ref{alg:SemiSupervisedLearning} to this semi-supervised learning problem. In particular, we attempt to reconstruct all the $70,000$ labels of the MINST data \cite{lecun1998gradient} from uniformly random sampled $35, 50, 70, 100$, and $700$ labels. For each sampling rate, we take 10 different random samples for comparisons. Figure \ref{fig:MINSTResult} shows the success rate of label estimation by graph Laplacian (GL) \cite{zhu2003semi}, weighted graph Laplacian (WGL) \cite{shi2016weighted}, and the proposed manifold based locally low-rank approximation based model (MLR). The first five images in Figure \ref{fig:MINSTResult} shows the success rate for each individual random sample with a fixed number of sample indices. The last image in Figure \ref{fig:MINSTResult} shows the average success rate which is naturally monotone increasing with respect to the number of sample indices. It can be clearly observed that the proposed method has the highest accuracy of estimation for almost all the random samples. In terms of average success rate, the proposed model outperforms the previously proposed graph Laplacian and weighted graph Laplacian based methods. We remark that further improvement can be expected if special treatments for shape recognition and similarity can be conducted which will be our future work.

\section{Conclusions}
\label{sec:conclusions}
In this paper, we propose a manifold based low-rank regularization method for image restoration and semi-supervised learning. The proposed regularization can be viewed as a point-wise linearization of the manifold dimension, which generalize the concept of low-rank regularization for linear objects as a concept of manifold based low-rank for nonlinear objects. Using the proposed regularization, we investigate new methods of image inpaining, image super-resolution and X-ray CT image reconstruction. We further extend this method to a general data analysis problem, semi-supervised learning. Intensive numerical experiments demonstrate that the proposed MLR method is comparable to or even outperforms the existing wavelet based models \cite{cai2008framelet,DongLiShen2012} and PDE based models \cite{zhu2003semi,shi2016weighted,osher2016low}.

Several directions will be investigated in our future work. For instance, the current method can be adapted to handle images with noisy input. It is also an important problem to explore a better method to pick the ``local regions" or manifold representation. For example, for semi-supervised learnings, the left image in Figure \ref{fig:MINSTLocalSimilar} shows that the KNN obtained by Euclidean distance may still include some ambiguity. In particular, some KNNs may have local rank as high as 7 or 8, which reduces the reliability of local low rank regularization. Therefore, developing a data-driven approach to non-Euclidean geometry for MLR will be a very interesting direction to investigate in our future work.

\section*{Acknowledgement} We thank Prof. Stanley Osher, Prof. Zuoqiang Shi and Mr. Wei Zhu kindly share their valuable comments and codes of both LDMM and LDMM+WGL for comparisons.


\end{document}